\documentclass[sigconf]{acmart}

\AtBeginDocument{%
  }

\setcopyright{acmcopyright}
\copyrightyear{2023}
\acmYear{2023}
\acmDOI{XXXXXXX.XXXXXXX}

\setcopyright{acmlicensed}
\acmConference[MM '23] {Proceedings of the 31st ACM International Conference on Multimedia}{October 29--November 3, 2023}{Ottawa, ON, Canada.}
\acmBooktitle{Proceedings of the 31st ACM International Conference on Multimedia (MM '23), October 29--November 3, 2023, Ottawa, ON, Canada}
\acmPrice{15.00}
\acmISBN{979-8-4007-0108-5/23/10}
\acmDOI{10.1145/XXXXXX.XXXXXX}

\usepackage{booktabs}
\usepackage{multirow}
\usepackage[normalem]{ulem}
\usepackage[switch]{lineno}
\useunder{\uline}{\ul}{}

\begin{document}
\nolinenumbers

\title{An Order-Complexity Aesthetic Assessment Model for Aesthetic-aware Music Recommendation}


\author{Xin Jin, Wu Zhou, Jinyu Wang}
\affiliation{%
  \institution{Department of Cyber Security, Beijing Electronic Science and Technology Institute, Beijing, 100070, China}
  \country{Beijing Institute for General Artificial Intelligence, Beijing, 100085, China}
}
\email{jinxin@besti.edu.cn}

\author{Duo Xu}
\authornote{Duo Xu and Yongsen Zheng are corresponding authors.}
\affiliation{%
  \institution{Tianjin Conservatory of Music, Tianjin, 300171, China}
  \country{Qingdao Academy of Intelligent Industries, Qingdao, 266114, China}
  \\
  \institution{Tongji University, Shanghai, 200082, China, many33@126.com}
}

\author{Yongsen Zheng}
\authornotemark[1]
\affiliation{%
  \institution{School of Computer Science and Engineering, Sun Yat-sen University, No. 132, East Outer Ring Road, Guangzhou-College-Town, Guangzhou, 510006, China}
  \country{z.yongsensmile@gmail.com}
}

\renewcommand{\shortauthors}{}


\begin{abstract}
    Computational aesthetic evaluation has made remarkable contribution to visual art works, but its application to music is still rare. Currently, subjective evaluation is still the most effective form of evaluating artistic works. However, subjective evaluation of artistic works will consume a lot of human and material resources. The popular AI generated content (AIGC) tasks nowadays have flooded all industries, and music is no exception. While compared to music produced by humans, AI generated music still sounds mechanical, monotonous, and lacks aesthetic appeal. Due to the lack of music datasets with rating annotations, we have to choose traditional aesthetic equations to objectively measure the beauty of music. In order to improve the quality of AI music generation and further guide computer music production, synthesis, recommendation and other tasks, we use Birkhoff’s aesthetic measure to design a aesthetic model, objectively measuring the aesthetic beauty of music, and form a recommendation list according to the aesthetic feeling of music. Experiments show that our objective aesthetic model and recommendation method are effective.
\end{abstract}



\maketitle

\section{Introduction}

    Computational aesthetic evaluation \cite{galanter2012computational, galanter2013computational} refers to using computational methods to assess the aesthetic beauty of different types of art, including paintings, photographs and music. It usually makes qualitative or quantitative evaluations \cite{jin2019aesthetic} to assess the quality of artworks generated by AI. It is necessary to consider the aesthetics of artworks because beauty often make people feel pleasant \cite{deng2017image}.
    
    \begin{figure}[htbp]
      \centering
      \includegraphics[width=0.5\textwidth]{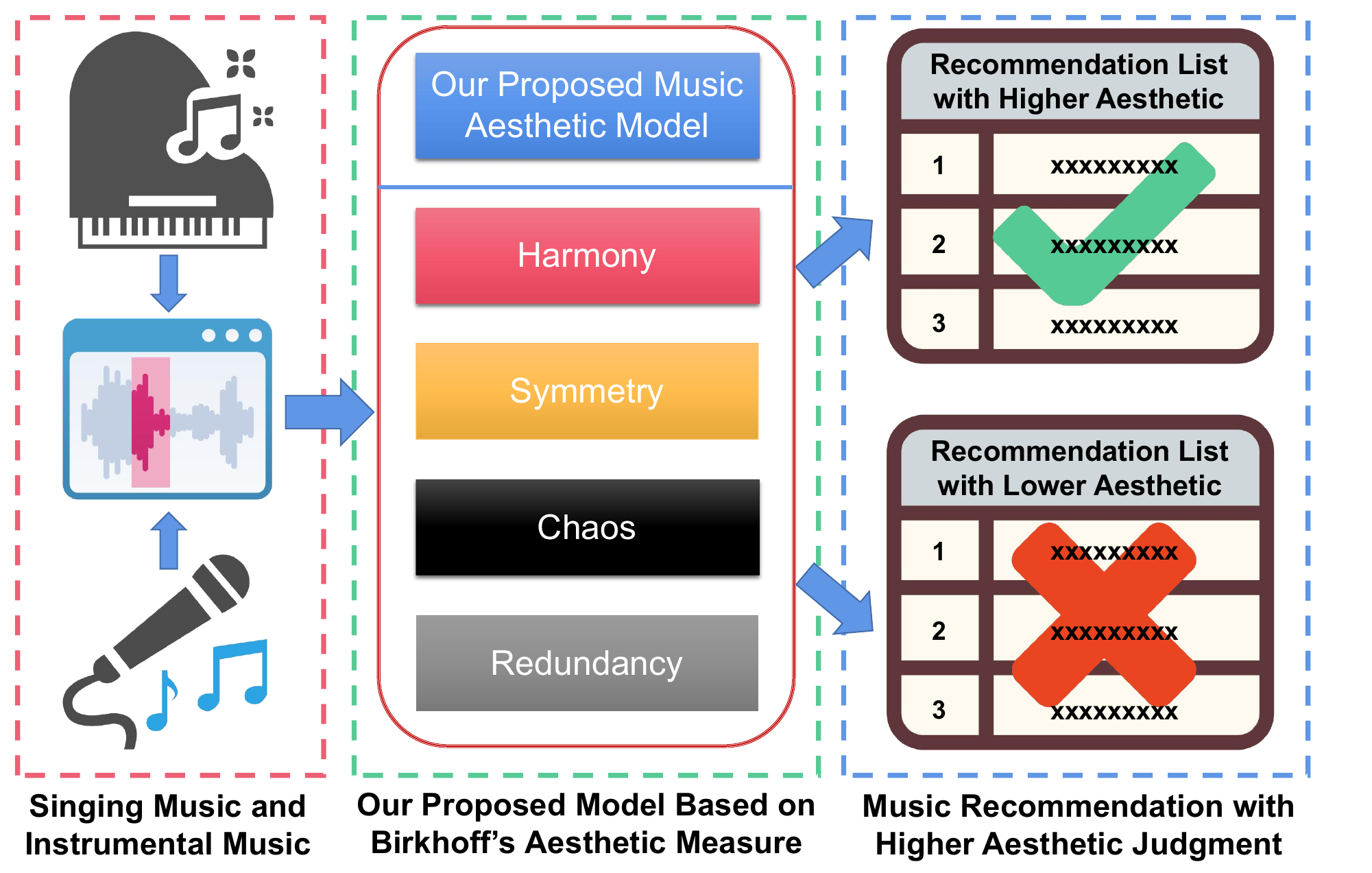}
      \caption{We construct a music recommendation list based on the results of the proposed music aesthetic model by adding four aesthetic features into music recommendation task.}
      \label{fig:introduction}
    \end{figure}
    
    In recent years, recommendation systems have flourished in various fields \cite{sun2019bert4rec}, including music related fields. Nowadays, when faced with such a massive amount of music resources, music recommendation can help users quickly find music of interest \cite{shen2020peia}. In fact, traditional recommendation systems often recommend music based on user preferences or the popularity of the music \cite{schedl2015music}, which may overlook some music works with higher artistic value.

    Generally speaking, the beauty of artistic works, especially the beauty of music, is highly abstract and subjective. The earliest research on music aesthetics is proposed by mathematicians (tracing back to Pythagoras) and attempt to solve it by abstracting it into emotional expression \cite{meyer1956meaning} or language \cite{lerdahl1996generative}. 
    
    We attempt to propose a method to objectively evaluate the beauty of music. Fig.\ref{fig:introduction} briefly shows our work.

\begin{figure}[htbp]
    \centering
      \includegraphics[width=0.5\textwidth]{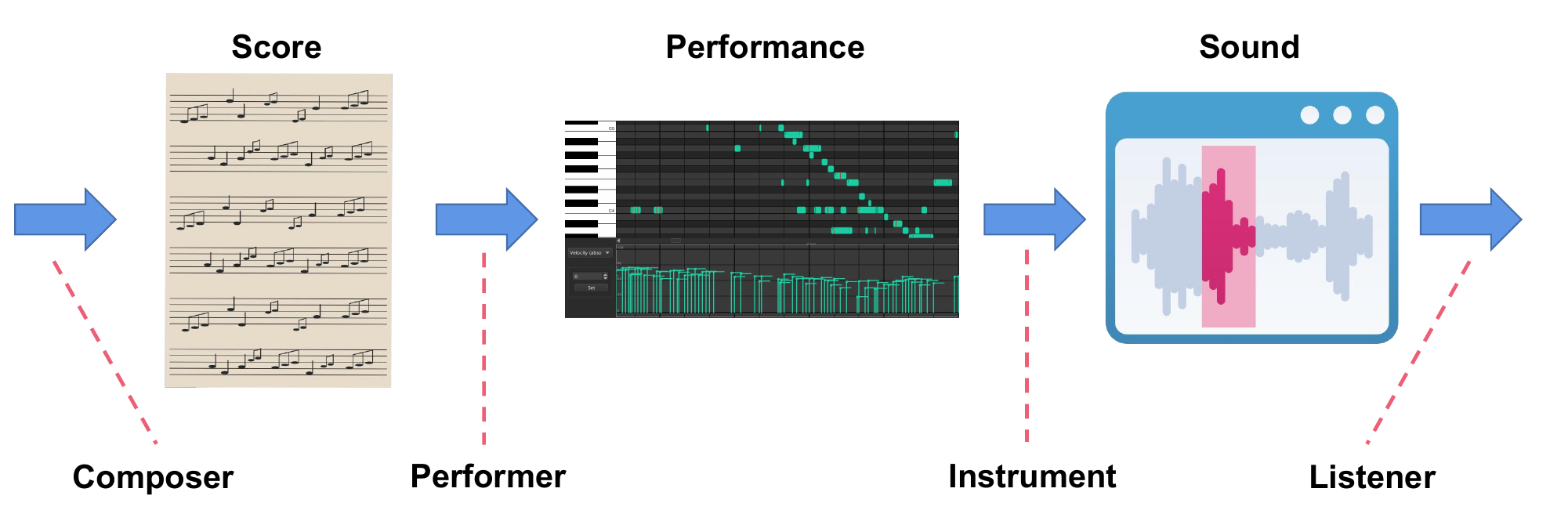}
      \caption{Three steps from creation to hearing for a music art work, and this paper focuses on the sound stage.}
      \label{fig:music}

    \end{figure}

    A musical art work usually needs to go through three steps from creation to hearing like Fig.\ref{fig:music}: Firstly, the musician composes and arranges music to make music score. Secondly, the performer plays music score to get performance. Finally, rendering the performance into sound according to different instrument selections \cite{oore2020time}.

    However, these music data are not labeled with aesthetic ratings like AVA \cite{murray2012ava} or AADB \cite{kong2016photo} in the field of image aesthetic evaluation. So we can't predict the beauty of music due to the lack of subjective labeled data. Due to the lack of datasets with subjective rating labels, we try to objectively measure the beauty of music. We adopt traditional measure to quantify the beauty of music.

    In this paper, Birkhoff’s method \cite{birkhoff2013aesthetic} was selected to conduct a study of aesthetic quality assessment of music. The reason why we choose Birkhoff's aesthetic measure is that it is not only famous, but also convenient to formalize, and has a relatively stronger interpretability for beauty (perhaps the most important reason). Birkhoff formalizes the aesthetic measure of an object into the quotient between order and complexity:

    \begin{equation}
        M=\frac{O}{C} \label{1}
    \end{equation}

    The taste and music creation of users are highly dependent on various factors, so traditional music recommendation systems often cannot meet users' needs and cannot provide good recommendations. Meeting users' music and entertainment needs requires considering their contextual \cite{baltrunas2011incarmusic} and good interactive information \cite{kulkarni2020context}. Therefore, what we focus on is that our recommendation method should not only make recommendations more accurate but also recommend more aesthetically pleasing music to users.

    The main contributions of our work are as follows:

    \begin{itemize}
        \item  We creatively put forward an objective method to measure the beauty of instrumental music and singing music.
        \item  We propose some basic music features and 4 aesthetic music features to facilitate the following computational aesthetic evaluation research tasks.
        \item  We apply the music aesthetics model to music recommendation tasks to recommend more aesthetically pleasing music to users, thereby enhancing their aesthetic taste.
    \end{itemize}

\section{Related Work}

\subsection{Music Generation}

    Music generation tasks is a critical research field of music information retrieval (MIR). The music generation task \cite{ji2020comprehensive} mainly includes three stages: score generation, performance generation and audio generation, and at present, the main methods of using deep learning to generate music include CNN \cite{oord2016wavenet, yang2017midinet}, RNN \cite{waite2016project, hadjeres2017deepbach, jeong2019virtuosonet}, VAE \cite{roberts2018hierarchical, brunner2018midi}, GAN \cite{dong2018musegan, huang2022singgan} and Transformer \cite{huang2018music, shih2022theme, zhang2022structure}.

    The contribution of WaveNet \cite{oord2016wavenet} in the field of audio generation is revolutionary, mainly based on its ability to generate high-quality audio waveforms, which can be used to generate audio and songs for various instruments. However, models such as WaveNet \cite{oord2016wavenet}, SampleRNN \cite{mehri2016samplernn}, and NSynth \cite{engel2017neural} are very complex, require a large number of parameters, a long training time, and a large number of examples, and lack interpretability. So other models considered, such as Dieleman et al. \cite{dieleman2018challenge} proposed the use of ADAs to capture long-term correlations in waveforms.

    The spectral characteristics of singing are also influenced by F0 motion \cite{yi2019singing}. Previously successful singing synthesizers are based on the concatenation method \cite{bonada2016expressive}. Although the systems are state-of-the-art in terms of sound quality and naturalness, their flexibility is limited and it is difficult to scale or significantly improve. The statistical parameter method \cite{oura2010recent} uses a small amount of training data for model adaptation. But these systems cannot match the existing smooth frequency and time.

\subsection{Music Evaluation}

    Objective evaluation is mainly used to assess the performance of music generation models. Objective evaluation metrics are designed to standardize the results of different models, allowing for comparison of their performance. These metrics are often statistical and can be categorized into four categories: pitch-related, rhythm-related, harmony-related, and style transfer-related \cite{ji2020comprehensive}, as used in models like MuseGAN \cite{dong2018musegan}. In addition, there are specialized metrics that do not directly relate to music, such as MV2H \cite{mcleod2018evaluating}, which evaluates the performance of music transcription models.

    Subjective evaluation is still the best way to assess aesthetics, and its methods include listening tests and visual analysis. Common listening tests include the Turing test \cite{hadjeres2017deepbach, cifka2019supervised} and side-by-side rating \cite{haque2018conditional}, which provide subjective evaluations of the music's quality and degree of innovation, as well as aspects such as harmony, rhythm, structure, coherence, and overall rating \cite{dong2018musegan}.
    
    There is currently limited research related to music aesthetics evaluation. SAAM \cite{jin2023order} is a state-of-the-art music aesthetics assessment model that provides aesthetic ratings for music scores. PAAM \cite{jin2023order2} is also a state-of-the-art music aesthetic model for evaluating the beauty of instrumental performances. Audio Oracle \cite{dubnov2011audio} uses Information Rate (IR) as an aesthetic measure.

    \begin{figure*}[htbp]
    \centering
      \includegraphics[width=\textwidth]{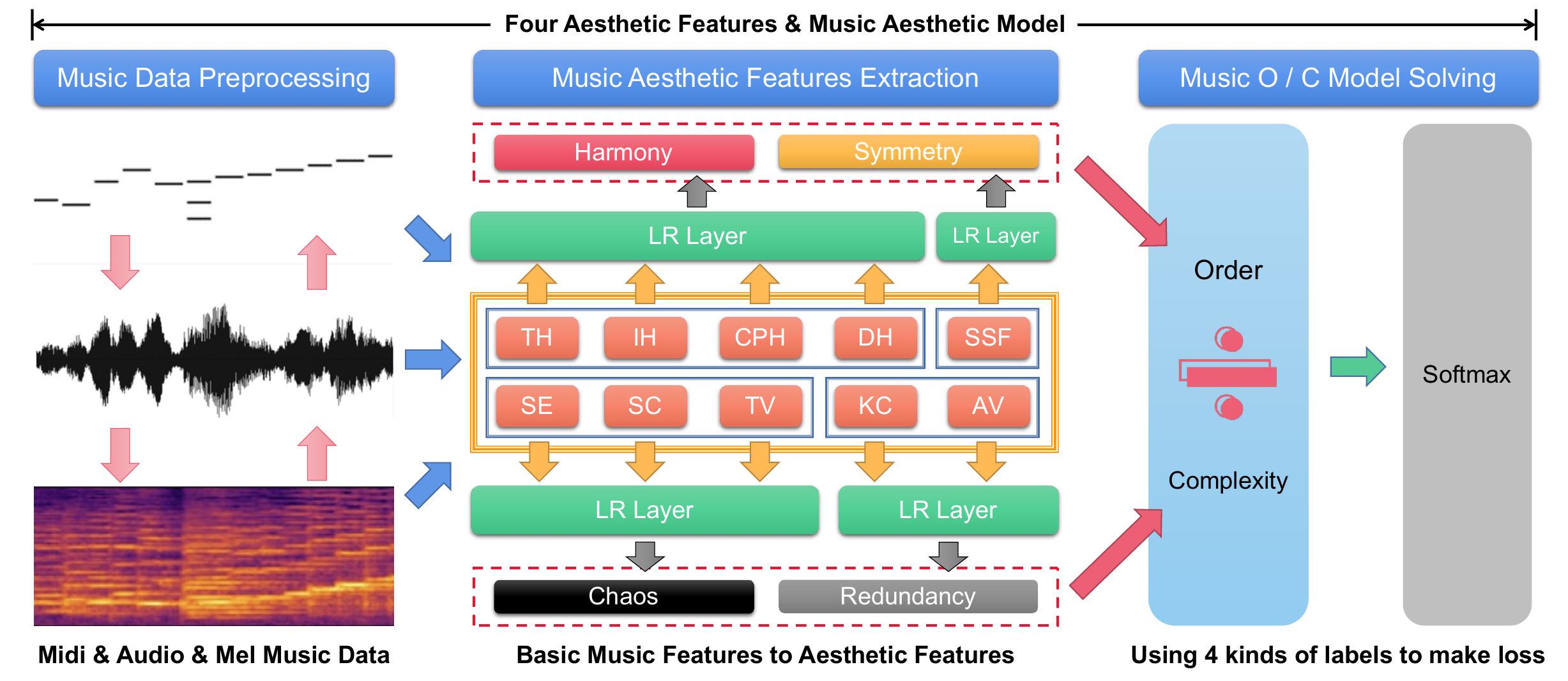}
      \caption{The overview of our aesthetic approach. We align and preprocess the midi, audio, and mel spectrogram of music, and obtain the values of ten basic music features by calculating their basic attributes. Then we put the corresponding basic music features to four logistic regression models to extract four music aesthetic features. Finally, we take the values of four music aesthetic features as the input of Birkhoff O/C model, and solve the model to get the final aesthetic score.}
      \label{fig:aesthetic}
    \end{figure*}
    
\subsection{Music Recommendation}

    Combining deep learning technology with traditional recommendation methods can effectively improve the effectiveness of music recommendation in recent years.
    
    Van et al. \cite{van2013deep} utilized the historical listening data of users and the audio signal data of music to project users and music into a shared hidden space, enabling them to learn the hidden representations of users and songs. Jiang et al. \cite{jiang2017play} proposed an improved algorithm based on RNN to measure the similarity between different songs. Wang et al. \cite{wang2014improving} proposed a DBN based method that combines the implicit representations of users and items into a user rating matrix for model training and improving the performance of music recommendations. Sachdeva et al. \cite{sachdeva2018attentive} proposed a novel attention neural structure that utilizes the features of these tracks to better understand users' short-term preferences.

\section{Music Aesthetic Model}

\subsection{Formalization}

    Based on Birkhoff's measure, we propose four aesthetic features: harmony, symmetry, chaos and redundancy. We linearly combine the order measures of molecules and the complexity measures of denominators by four logistic regression model. Fig.\ref{fig:aesthetic} shows our method. Detailed measures explanation will be described for next. The music aesthetic measure formalization is as follows:

    \begin{equation}
        Aesthetic\ Measure=\frac{\omega_1H + \omega_2S + \theta_1}{\omega_3C + \omega_4R + \theta_2} \label{6}
    \end{equation}

    Where $H$ is harmony, $S$ is symmetry, $C$ is chaos and $R$ is redundancy. $\omega$ is the weight and $\theta$ is the constant.

    \subsection{Harmony}
    
    People feel good because there is “harmony” in music \cite{clemente2022musical}.
    
    \subsubsection{Timbre Harmony}
    
The formalization of timbre harmony can be divided into two aspects: timbre balance and complementarity.

Timbre balance refers to the relative balance of sound energy in different frequency bands among instruments.

Let $S_i(f)$ denote the sound energy of the $i$-th instrument in frequency band $f$, and let $F$ be the set of all frequency bands. Then, timbre balance can be represented as:

    \begin{equation}
        B = \max_{f \in F} \left(\frac{\max_{i=1}^n S_i(f)}{\min_{i=1}^n S_i(f)}\right)
    \end{equation}
    
Where $n$ is the number of instruments, $B$ represents the ratio of the minimum and maximum sound energy proportions among different frequency bands, and $B \in [0, 1]$.

Timbre complementarity refers to the mutual complementation of sound colors among instruments.

Let $s_i(t)$ denote the sound waveform of the $i$-th instrument at time $t$, and let $s(t)$ denote the total sound waveform of the ensemble. Then, timbre complementarity can be represented as:
    \begin{equation}
        C = \frac{1}{T}\int_0^T\sum_{i=1}^n s_i(t)s(t)dt
    \end{equation}

Where $T$ is the total length of the music, and $C$ represents the timbre complementarity score among instruments, with $C \in [-1, 1]$.

\begin{equation}
Timbre \ Harmony = \alpha \cdot B + (1-\alpha) \cdot C
\end{equation}

Timbre harmony is obtained by linearly weighting balance and complementarity. Where $\alpha$ is the weighting factor.

    \subsubsection{Interval Harmony}

    The term "interval" refers to the distance between two notes in music. An interval of 12 semitones is called an octave. Mathematical and physical studies have demonstrated that when two sound frequencies have a simple integer ratio, they produce a more pleasing auditory experience when played together. To measure the harmony of intervals, We refer to the SAAM \cite{jin2023order} method and propose the following formula.

    \begin{equation}
    Interval\ Harmony=\sum_{i=1}^{12} \alpha_i * pir_i + \theta_{ih} \label{1}
    \end{equation}
    
    In this formula, $\alpha_i$ represents the weight of each interval, $pir_i$ is the ratio of the specific interval to the total number of intervals, and $\theta_{ih}$ is a constant term.

    \subsubsection{Chord Progression Harmony}

    In tonal music, harmony progression is a specific harmonic range in chords connection.
    
    In this study, we adopt the method in SAAM \cite{jin2023order}, which calculates the average value of progression tension to obtain a quantitative chord progress harmony. The formula is as follows:
    
    \begin{equation}
    \begin{split}
    Chord\ Progression\ Harmony = \lambda_1 d_1(T_i,T_{i-1})+ \\
    \lambda_2 d_2(T_i, T_{key}) + \lambda_3 d_3(T_i-T_{key}, T_f) + \lambda_4 c(T_i) + \\
    \lambda_5 m(T_i,P) + \lambda_6 h(T_i,P)
    \end{split}
    \end{equation}
    
    Here, $T_i$ represents the i-th chord of the progression $P$, and $\lambda$ is the weight parameter. Further details on the parameters $c$, $m$, and $h$ can be found in \cite{navarro2020computational}.

    \subsubsection{Dynamic Harmony}
    
    According to GTTM theory \cite{lerdahl1996generative}, the metric structure of a piece of music can be derived from it. The strength of playing each note can be described as dynamic. The metrical structure of music represents the strong and weak beats.
    
    We refer to the PAAM \cite{jin2023order2} approach to quantify the harmony of dynamics. The formula for dynamic harmony is:
    
    \begin{equation}
    Dynamic\ Harmony = \frac{\sum_{i=1}^{n} D_i \times M_i}{\sqrt{\sum_{i=1}^{n} (D_i)^2} \times \sqrt{\sum_{i=1}^{n} (M_i)^2}}
    \end{equation}
    
    Here, $i$ represents the $i$-th note at the beginning of the bar, $n$ is the total number of aligned notes at the beginning of the bar, $D$ represents the vector of dynamic values, and $M$ represents the vector of metrical structure values.

    \subsection{Symmetry}
    
    Symmetry is considered to be the key to the perfection of art \cite{osborne1986symmetry}.
    
    \subsubsection{Self-Similarity-Fitness}

     In music, symmetry usually refers to a form of regularity, repetition, or mirroring in the music. In the music generation field, the structure of a piece is often regarded as a crucial aspect. Repetitive patterns can be found in almost all kinds of music. Our aim is to find how the repetitive structure affects the beauty of music.

    Drawing inspiration from the aesthetics of art and images \cite{al2017symmetry}, we hypothesize that the beauty of music originates from the symmetry present in its compositions. Thus, we adopt Müller's fitness method \cite{MuellerJG13_StructureAnaylsis_IEEE-TASLP} to gauge the extent of repetition in a musical piece. The fitness formula is expressed as follows:
    
    \begin{equation}
    Self\ Similarity\ Fitness = 2 \cdot \frac{\bar{\sigma}(\alpha) \cdot \bar{\gamma}(\alpha)}{\bar{\sigma}(\alpha)+\bar{\gamma}(\alpha)}
    \end{equation}

    Where $\bar{\sigma}(\alpha)$ and $\bar{\gamma}(\alpha)$ refer to Müller's method \cite{MuellerJG13_StructureAnaylsis_IEEE-TASLP}.


    \subsection{Chaos}
       When people deal with complicated information, they will feel uncomfortable and uneasy \cite{bense1960programmierung}. 
    \subsubsection{Shannon Entropy}

    We employ Shannon entropy to quantify the degree of chaos or disorder in the internal state of a system, using histogram entropy as a specific instance of this measure. Given a finite set $\Omega$ and a random variable $X$ taking values in $\Omega$, with each value $x\in\Omega$ having a probability distribution $p(x) = Pr[X=x]$. What we need to consider is the entropy of the music attribute histogram. The Shannon entropy $H(X)$ is defined as:

        \begin{equation}
            Shannon \ Entropy=-\sum_{x\in\Omega}p(x)\log(x)
        \end{equation}

    The Shannon entropy is a widely used method to assess the average uncertainty of a random variable $X$.

    \subsubsection{Spectral Complexity}

    Spectral complexity is a method for measuring the complexity of a musical signal's spectral content. It is based on the spectral flux of the signal. It is a measure of how much the spectral content of a signal changes over time.

    The spectral complexity of the signal is then defined as the weighted sum of the spectral flux:

    \begin{equation}
        Spectral \ Complexity = \frac{1}{T}\sum_{t=1}^{T} \sum_{f=1}^{F} |X(f, t) - X(f, t-1)|
    \end{equation}

    Where $x(t)$ is a signal with $N$ samples, $X(f, t)$ is its short-time Fourier transform, $F$ is the number of frequency bins, $T$ is the number of analysis frames.

    \subsubsection{Timbre Variability}

    Timbre variability is a method used to measure the diversity of timbre in music. It can calculate the degree of timbre variation in the music.

    Formally, let $X={x_1,x_2,...,x_n}$ represent the $n$ notes in the music, $f(x_i)$ represent the basic timbre feature of the $i$th note, and $Var(x_i)$ represent the timbre variability of the $i$th note. Then the Timbre variability of the entire piece of music can be represented as:
        \begin{equation}
    Timbre \ Variability =\frac{\sum_{i=1}^n Var(x_i)\cdot w_i}{\sum_{i=1}^n w_i}
        \end{equation}
        
    Here, $w_i$ represents the weight of the $i$th note.


    
    \subsection{Redundancy}

    Aesthetically speaking, redundancy makes people feel dull, resulting in negative emotions \cite{lorand2002aesthetic}.

    \subsubsection{Kolmogorov Complexity}

    The Kolmogorov complexity $K(s)$ of a string $s$ is defined as the length of the shortest program that outputs $s$ on a computer. In the context of music, lossless compression can be used to estimate the redundancy of a piece of music. This can be expressed using the following formula:

    \begin{equation}
        Kolmogorov\ Complexity=1 - \frac{K}{I_m}
    \end{equation}

    Where $K$ is the amount of information after lossless compression of music, and $I_m$ is the original amount of information of music.
    
    \subsubsection{Autocorrelation Value}

    The autocorrelation value reflects the degree of redundancy in music.

    Given a music signal $x(t)$, the autocorrelation value can be calculated as:
        \begin{equation}
    Autocorrelation \ Value = \frac{1}{N}\sum_{i=0}^{N-1}\frac{\sum_{t=0}^{T-i-1} x(t)x(t+i)}{\sum_{t=0}^{T-1}x^2(t)}
        \end{equation}
        
    Where $N$ is the maximum time lag for calculating the autocorrelation, $T$ is the length of the music signal, and $x(t)$ is the amplitude of the signal at time $t$.

        \begin{figure*}[htbp]
    \centering
      \includegraphics[width=\textwidth]{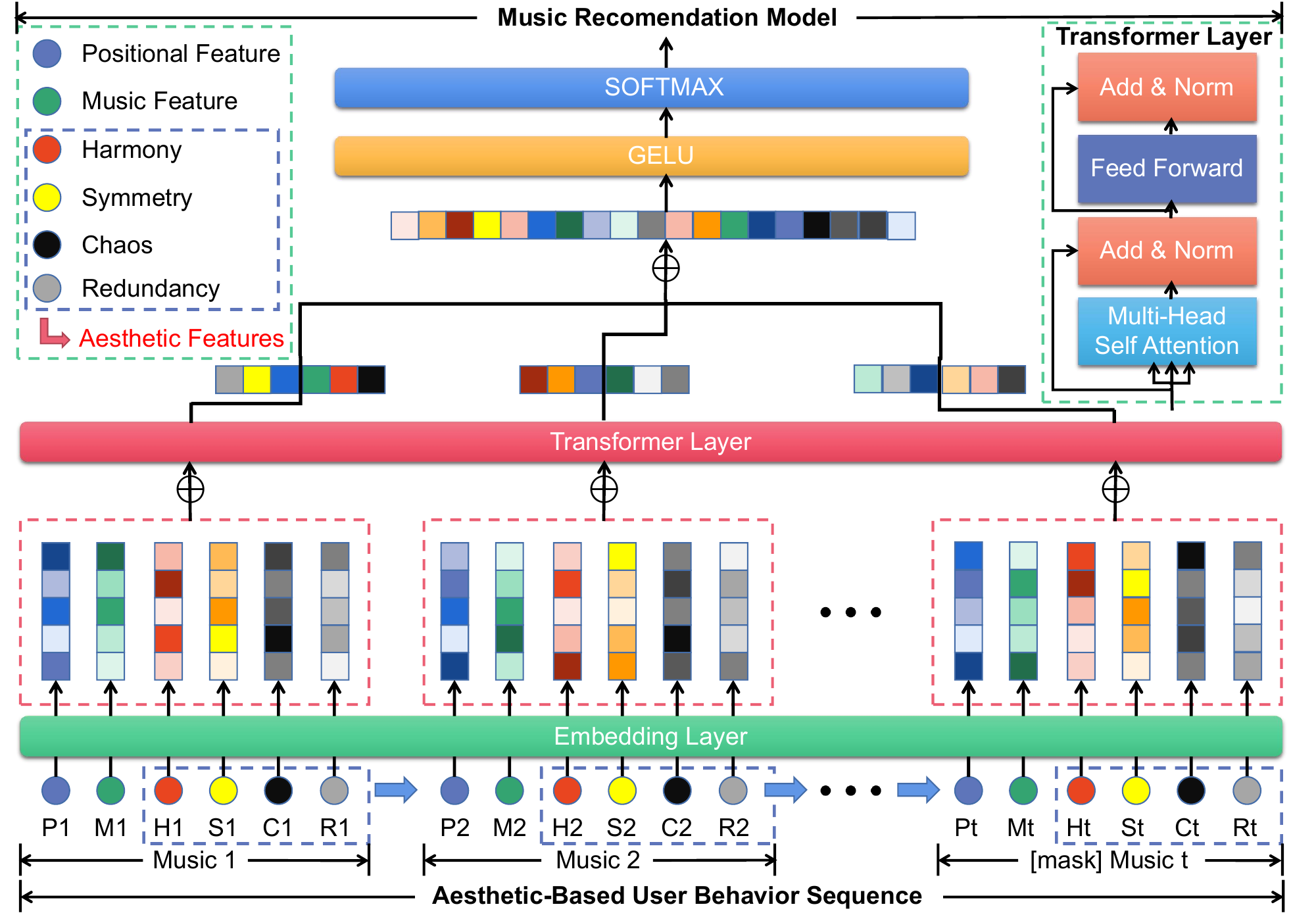}
      \caption{The overview of our recommendation approach. In addition to traditional positional features and music features, we will also extract good aesthetic features as important reference standards for sequential music recommendations. Moreover, we also consider the aesthetic rating score of music as a label value in the recommendation of sequential music.}
      \label{fig:recommendation}
    \end{figure*}
    \section{Aesthetic-aware Recommendation}


    
We use CL4SRec \cite{xie2022contrastive} as the backbone network of our method, which is the state-of-art model for sequence recommendation tasks.
\subsection{Embedding Layer}

\subsubsection{Positional Embedding} We transform positional features into vector representations. We assign a unique number to each position in the sequence and map each of these to a d-dimensional vector space to obtain position's vector representation.
    
Specifically, we use a matrix $E_{pos}$ of size $L \times d$ to represent the vector representations of all positions, where $E_{pos_i}$ represents the vector representation for position $i$. We map the position identifier to the $i$-th row of this matrix, $E_{pos_i}$.

Then, we obtain the representation for each position $\mathbf{X}_{pos}$. Then we will discuss the embedding methods for other features.

\subsubsection{Other Features Embedding} For the music features and the four aesthetic features, we can use corresponding embedding layers to convert them into fixed-length vector representations.

Specifically, we use an embedding matrix $E_{music}$ of size $D_{music}$ to represent the vector representation of music features (CNN), where $D_{music}$ is the embedding dimension. For each music feature $j$, we represent it as a vector $\mathbf{v}j \in \mathbb{R}^{D{music}}$, and then map it to a row of the embedding matrix $E_{music}$, i.e., $E_{music_j} = \mathbf{v}_j$.

Similarly, we use four embedding matrices $E_{aes}$ of size $D_{aes}$ to represent the vector representation of aesthetic features.

Then, we obtain the representation for music feature $\mathbf{X}_{music}$ and aesthetic features $\mathbf{X}{aes}$.

\subsubsection{Embedding Overview}

We add these vector representations with the position embedding vector representations to form the input matrix $\mathbf{X}$ of the Transformer model:

\begin{equation}
    \mathbf{X} = \mathbf{X}^{(0)} + \mathbf{X}{pos} + \mathbf{X}{music} + \mathbf{X}{aes}
\end{equation}

Where $\mathbf{X}^{(0)}$ represents the vector representation for the original input sequence, $\mathbf{X}{pos}$ is the position embedding matrix, and $\mathbf{X}{music}$ and $\mathbf{X}{aes}$ are the embedding matrices for music and aesthetic features, respectively.

\subsection{Transformer Layer}

\subsubsection{Multi-Head Self-Attention} For each attention head $h$, we compute the attention weights $\mathbf{A}^h$ as follows:

\begin{equation}
    \mathbf{A^{h}} = \textbf{softmax}\left(\frac{QW_{i}^{h}(KW_{i}^{h})^{T}}{\sqrt{d_k}}\right)
\end{equation}

Where $\mathbf{Q}, \mathbf{K}, \mathbf{V} \in \mathbb{R}^{n \times d_k}$ are the query, key, and value matrices obtained by linearly projecting the input sequence $\mathbf{X}$ with learned weight matrices $\mathbf{W}_q, \mathbf{W}_k, \mathbf{W}v \in \mathbb{R}^{d \times d_k}$.

The output of the Multi-Head Self-Attention layer is then obtained by concatenating the $h$ attention heads and passing the result through a linear layer:

\begin{equation}
    \textbf{MultiHead(Q, K, V)}= \text{Concat}(\text{head}_1, \text{head}_2, ..., \text{head}_h) \textbf{W}_o
\end{equation}

\begin{equation}
    \text{head}_i = \textbf{Attention}(Q\mathbf{W}_i^q, K\mathbf{W}_i^k, V\mathbf{W}_i^v)
\end{equation}

Where $\mathbf{W}o \in \mathbb{R}^{hd_v \times d}$ is a learned weight matrix.

\subsubsection{Position-wise Feed-Forward Network} The feed-forward neural network consists of two fully connected layers with an activation function applied after each layer. It can be written as:

\begin{equation}
    \mathbf{FFN(x)}=\mathbf{GELU}(\mathbf{x}\mathbf{W}_1+\mathbf{b}_1)\mathbf{W}_2+\mathbf{b}_2
\end{equation}

\begin{equation}
    \mathbf{GELU(x)} = x \Phi(x)
\end{equation}

Where $\mathbf{W_1}$ and $\mathbf{W_2}$ are both learnable weight matrices, $\mathbf{b}1$ and $\mathbf{b}2$ are the bias terms for the first and second layers, respectively.

\subsubsection{Stacking Transformer Layer} The layer consists of $L$ identical layers. Each layer applies a multi-head self-attention mechanism followed by a position-wise feed-forward network (FFN) to the input sequence. Additionally, each layer employs residual connections and layer normalization to stabilize the training process.

The output of the $l$-th layer is given by:

\begin{equation}    
\begin{split}
    \mathbf{X_l} = \textbf{LayerNorm}\big(\mathbf{X_{l-1}} + \textbf{MultiHead}\big(\textbf{LayerNorm}\big(\mathbf{X_{l-1}}\big)\big)\big) + \\ \textbf{FFN}\big(\textbf{LayerNorm}\big(\mathbf{X_{l-1}} + \textbf{MultiHead}\big(\textbf{LayerNorm}\big(\mathbf{X_{l-1}}\big)\big)\big)\big)
\end{split}
\end{equation}

Where $l \in {1, 2, ..., L}$ indicates the $l$-th layer of the stacking transformer layer. $\text{LayerNorm}(\cdot)$ represents the layer normalization operation, $\text{MultiHead}(\cdot)$ represents the multi-head self-attention, and $\text{FFN}(\cdot)$ represents the position-wise feed-forward network.

\subsection{Output Layer}
For the music sequence recommendation task, the output layer can be formulated as follows: given the final output $H_L$ from the L layers of the hierarchical exchange information, we mask the item $v_t$ at time step $t$ and then use $h_{Lt}$ to predict the masked item $v_t$. We apply a two-layer feed-forward network with GELU activation in between to produce an output distribution over target items:

\begin{equation}
    \mathbf{P(v_t)} = \textbf{softmax} \left(W_P \cdot \text{GELU}\left(h_{Lt} \cdot E\right) + b_P\right) + b_O
\end{equation}

Where $\mathbf{W}_P$ is the learnable projection matrix, $\mathbf{b}_P$ and $\mathbf{b}_O$ are bias terms, and $\mathbf{E} \in \mathbb{R}^{|V| \times d}$ is the embedding matrix for the item set $V$.
\subsection{Model Learning}
We train the model using the randomly masked language modeling (LM) technique, where we randomly mask portions of song IDs and train the model to predict the original value of the masked song IDs using the context about non-masked song IDs in the sequence:

\begin{equation}
    Input \ layer[\text{s1,s2,s3,s4,s5}] \rightarrow [\text{s1,s2, [mask1],s4, [mask2]}]
\end{equation}

The cross-entropy function can be represented as follows:

\begin{equation}
    J=-\frac{1}{N} \sum_{i=1}^{N} \log P(v_{ti}|v_{1i},...,v_{t-1,i},v_{t+1,i},...,v_{ni})
\end{equation}

Here, $N$ represents the number of training samples, and $v_{ti}$ represents the masked music item in the $i$-th training sample.

\section{Experiments}

\subsection{Datasets \& Setup}

\subsubsection{Dataset For Aesthetic Model} The dataset is divided into three types of samples: real audio played or sung by humans (positive), performance midi rendered audio (medium), and audio generated by various AI models (negative), 7:3 for training and testing. 

Specifically, our positive sample uses MedleyDB \cite{Bittner2014}, our mid value is audio rendered (by MuseScore3) from POP909's \cite{wang2020pop909} piano and Lakh MIDI Dataset's \cite{raffel2016learning} vocal performance midi, and the negative sample uses Music Transformer \cite{huangmusic} to generate performance midi, and then use GanSynth \cite{engel2019gansynth} to synthesize it into audio.

    \begin{table}[htbp]
    \begin{tabular}{ccc}
    \toprule
    \textbf{AI (negative)} & \textbf{Rendered (medium)} & \textbf{Human (positive)} \\ \midrule
    122            & 140                       & 128           \\ \bottomrule
    \end{tabular}
    \vspace{0.3cm}
            \caption{The number of samples in the aesthetic model.}
            \label{tab:tab1}
    \end{table}
\vspace{-1cm}
\subsubsection{Recommendation Datasets} We select three datasets:

    \begin{itemize}
        \item  MSD \cite{Bertin-Mahieux2011} dataset is a collection of metadata and feature analysis data for one million songs, including information about artists, albums, tracks, lyrics, genres, tonality, and more.
        \item  Last. fm-1k \cite{Celma2008} dataset contains implicit feedback data from users on music, including music play records, music tags, user information, and more.
        \item  Last. fm-360k \cite{Cantador2011} dataset adds user-artist relationship and user information on top of the 1k dataset.
    \end{itemize}

    We clean the dataset by removing playback events from the session information that lack user or information, and only keeping session information with playback events greater than 10. Then we crawl corresponding audio source files according to playlists, and use Onsets and Frames \cite{hawthorne2017onsets} to get the corresponding MIDI.

        \begin{figure*}[htbp]
    \centering
      \includegraphics[width=\textwidth]{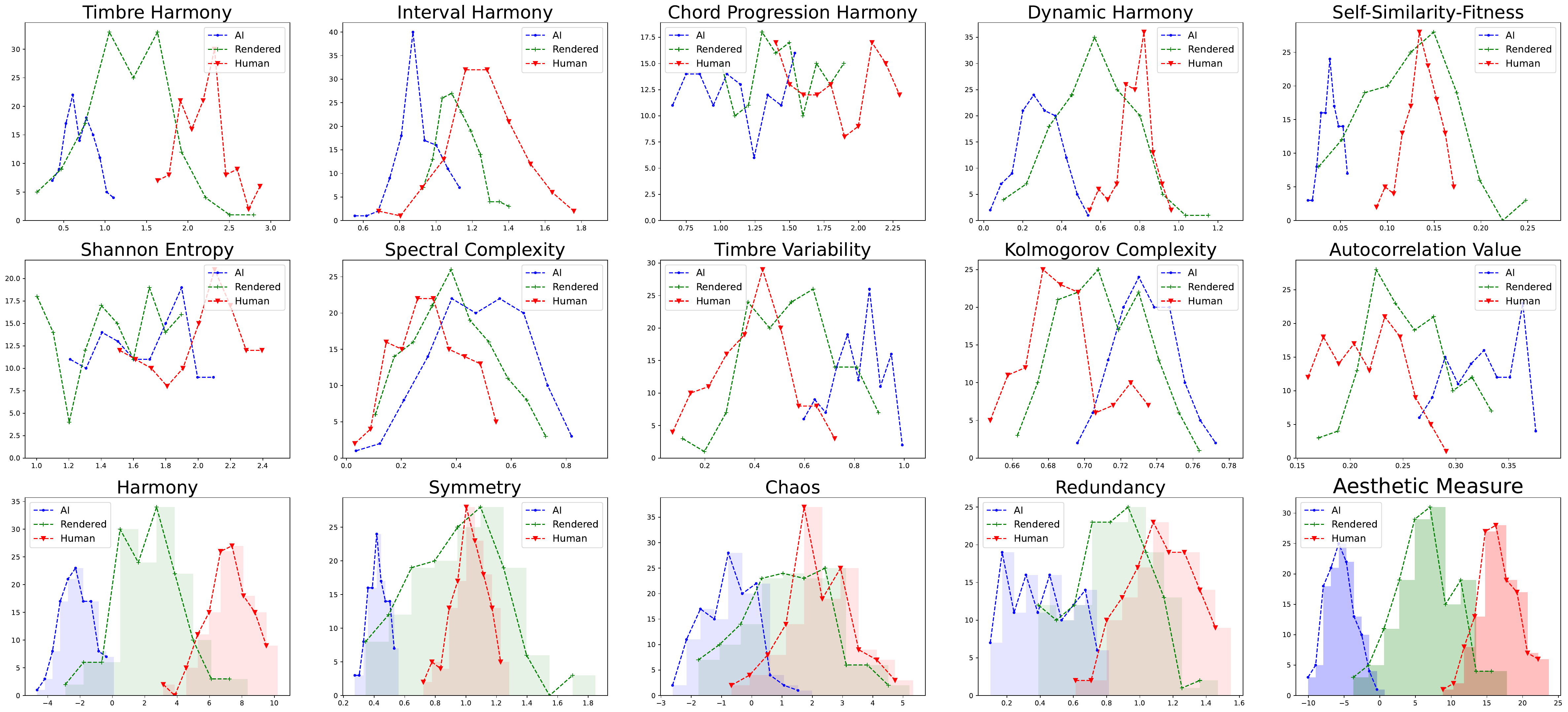}
      \caption{The first two rows in the figure show the distribution of ten basic music features on three sample sets. The last graph in the third row shows the aesthetic score distribution of our trained model on the three sample sets. The remaining four graphs in the third row show the distribution of four aesthetic features on the three sample sets.}
      \label{fig:afd}
    \end{figure*}

\subsubsection{Evaluation Metrics}

    For the aesthetic model scoring task, we evaluate the performance of the classification model using classification accuracy and F1 measure. For recommended tasks, we use evaluation metrics to assess the ranking list generated by the models, which include Hit Ratio (HR), Normalized Discounted Cumulative Gain (NDCG). As we only have one true item for each user, HR@k is equivalent to Recall@k and proportional to Precision@k. Additionally, we report HR and NDCG at k = 1, 5, and 10, where a higher value indicates better performance.

\subsection{Implementation Details}

\subsubsection{Music Aesthetic Model}

    We use music21\footnotemark{} \footnotetext{https://github.com/cuthbertLab/music21} to obtain music attributes, Musescore3\footnotemark{} \footnotetext{https://musescore.org/} to render music, and jSymbolic \cite{mckay2018jsymbolic}, jAudio \cite{mckay2005jaudio}, and librosa \cite{mcfee2015librosa} to extract corresponding music features to complete the calculation of basic music features. For calculation, we use the aligned midi, audio, and mel. Regarding the four aesthetic features, we will linearly combine the corresponding basic music features and use four logistic regression models for pre training to obtain aesthetic features. Finally, input the value of aesthetic features into the O/C model, predict the label, select the cross entropy as the loss function, and use the Adam \cite{kingma2014adam} optimizer for gradient descent. We set the learning rate to 5e-5, and after 1000 iterations, the loss function converges. For more details, please refer to the implementation of SAAM \cite{jin2023order} and PAAM \cite{jin2023order2}.

\subsubsection{Music Recommendation Model}

    We use word2vec to process positional features, CNN to process music features, and for four aesthetic features, we use the original vector. We train the model using TensorFlow with truncated normal distribution initialization in the range of [-0.01, 0.01] for all parameters. The optimization is performed using the Adam \cite{kingma2014adam} with a learning rate of 1e-5, $\beta_1 = 0.9$, $\beta_2 = 0.999$, $\ell_2$ weight decay of 0.01, and linear decay of the learning rate. The gradient is clipped when its $\ell_2$ norm exceeded a threshold of 5. To ensure fair comparison, we set the layer number $L = 2$ and head number $h = 4$. We select the mask proportion $\rho$ through validation, resulting in $\rho$ = 0.6 for MSD, and $\rho$ = 0.4 for Last.fm-1k and Last.fm-360k. Model is trained on NVIDIA GeForce GTX 1080Ti GPU with a batch size of 256.

\subsection{Results Analysis \& Discussion}

\subsubsection{Music Aesthetic Model}

    The distribution of various indicators and the final aesthetic measure in the aesthetic model are shown in Fig.\ref{fig:afd}. After training our O/C model, we observe the intersection of the aesthetic scores distributions (in the last of the third row in Fig. 5) is very small, indicating the O/C model's necessity. To be more specific, TH, IH, and DH have achieved good differentiation effects, proving that human music audio is significantly superior to rendered audio and AI generated audio in terms of harmony. As for the symmetry aspect, the audio generated by AI has a significantly lower repeatability and appears to be structurally disorganized, while the differentiation between the other two is not as significant. In terms of Chaos, the differentiation of basic music features is not particularly significant. In terms of redundancy, it is clear that human audio is the best, followed by the other two. The accuracy of our model on the test set is 92.3\%, and the f1 measure is 90.5\%.
    
\vspace{-1cm}
\subsubsection{Music Recommendation Model}
    In Table\ref{tab:tab2}, we present a summary of the best performance achieved by all models on four benchmark datasets. As the NDCG@1 results are equivalent to HR@1 in our experiments, we have omitted them from the table. We have also applied some other networks (7 baseline models) as backbone networks, namely BPR \cite{rendle2012bpr}, CDAE \cite{wu2016collaborative}, FPMC \cite{rendle2010factorizing}, GRU4Rec \cite{hidasi2015session}, Caser \cite{tang2018personalized}, SASRec \cite{kang2018self}, Bert4Rec \cite{sun2019bert4rec} and the performance comparison will be provided in Table\ref{tab:tab2}. The experimental results show that the performance is best when using CL4SRec as the backbone network, CL4SRec is superior in nearly all metrics. Note that in the results of Table\ref{tab:tab2}, we only add aes features on CL4SRec. The penultimate line is CL4SRec without aesthetic features, and our performance has slightly improved after integrating aesthetic features, proving the aesthetic features are meaningful.

\begin{table*}[htbp]
\centering
\tabcolsep=0.23cm
\begin{tabular}{@{}llccccccccc@{}}
\toprule
Datasets                      & Metric  & BPR    & CDAE   & FPMC  & GRU4Rec   & Caser &  SASRec       &  BERT4Rec      & CL4SRec          & CL4SRec w/ aes\\ \midrule
\multirow{5}{*}{MSD}          & HR@1    & 0.0465 & 0.0479 & 0.0474 & 0.0493 & 0.0577  & 0.0703       & 0.0714 & \textbf{0.0936} & {\ul 0.0927}\\
                              & HR@5    & 0.1182 & 0.1248 & 0.1242 & 0.1319 & 0.1681  & 0.1692       & 0.1828 & {\ul 0.2179}  & \textbf{0.2185}\\
                              & HR@10   & 0.1978 & 0.2014 & 0.2148 & 0.2369 & 0.2654  & 0.2687 & 0.2584       & {\ul 0.3071}  & \textbf{0.3089}\\
                              & NDCG@5  & 0.0941 & 0.0754 & 0.0831 & 0.0768 & 0.1146  & 0.1195       & 0.1282 & {\ul 0.1583} & \textbf{0.1624}\\
                              & NDCG@10 & 0.1023 & 0.1273 & 0.1252 & 0.1342 & 0.1564  & 0.1532       & 0.1591 & \textbf{0.1902} & {\ul 0.1895}\\ \midrule
\multirow{5}{*}{Last.fm-1k}   & HR@1    & 0.0443 & 0.0517 & 0.0392 & 0.0374 & 0.0633  & 0.0669       & 0.0704 & {\ul 0.0982} & \textbf{0.1014}\\
                              & HR@5    & 0.1138 & 0.1532 & 0.1309 & 0.1414 & 0.1795  & 0.1863       & 0.2384 & {\ul 0.2709} & \textbf{0.2718}\\
                              & HR@10   & 0.2042 & 0.2251 & 0.2389 & 0.2446 & 0.2647  & 0.2986       & 0.3613 & {\ul 0.3999} & \textbf{0.4026}\\
                              & NDCG@5  & 0.1045 & 0.0985 & 0.1079 & 0.0963 & 0.1408  & 0.1675 & 0.1501       & {\ul 0.1821} & \textbf{0.1902}\\
                              & NDCG@10 & 0.1152 & 0.1203 & 0.1269 & 0.1291 & 0.1832  & 0.1859       & 0.1952 & \textbf{0.2254} & {\ul 0.2202}\\ \midrule
\multirow{5}{*}{Last.fm-360k} & HR@1    & 0.0768 & 0.1248 & 0.1275 & 0.1481 & 0.1535  & 0.1858       & 0.2151 & {\ul 0.2854} & \textbf{0.2936}\\
                              & HR@5    & 0.2984 & 0.3095 & 0.3730 & 0.4118 & 0.4528  & 0.4890       & 0.5272 & {\ul 0.5894} & \textbf{0.5931}\\
                              & HR@10   & 0.4109 & 0.4627 & 0.5219 & 0.5809 & 0.5976  & 0.6012       & 0.6467 & {\ul 0.6972} & \textbf{0.6981}\\
                              & NDCG@5  & 0.1879 & 0.2173 & 0.2418 & 0.3012 & 0.3623  & 0.3471       & 0.3767 & \textbf{0.4484} & {\ul 0.4313}\\
                              & NDCG@10 & 0.2164 & 0.2579 & 0.2841 & 0.3448 & 0.3687  & 0.3512       & 0.4176 & {\ul 0.4807} & \textbf{0.4984}\\ \bottomrule
\end{tabular}
\vspace{0.15cm}
\caption{The performance of various methods on next-item prediction is compared, with the best scores in each row shown in bold and the second-best scores underlined. Note that, we only add aes features on CL4SRec, which achieves best performance.}
\label{tab:tab2}
\vspace{-1.5em}
\end{table*}

\subsection{Subjective Evaluation}

We refer to this work \cite{dong2022deep}. To confirm the effectiveness of our aesthetic model and recommendation algorithm, we conduct a subjective evaluation experiment and invite 15 professional musicians to participate. It is necessary to confirm the validity of the dataset, that is, whether the audio generated by AI has the lowest aesthetic value, followed by rendered audio, and the audio actually performed or sung by humans has the highest aesthetic value. We select 5 songs from each of the above three sample sets and ask volunteers to rate their aesthetic value. The results are shown in Table\ref{tab:tab3}:

\begin{table}[htbp]
\centering
\tabcolsep=0.25cm
\begin{tabular}{@{}ccc@{}}
\toprule
\textbf{AI (negative)} & \textbf{Rendered (medium)} & \textbf{Human (positive)} \\ \midrule
2.32 ± 0.27  & 3.25 ± 0.32   & 4.18 ± 0.16     \\ \bottomrule
\end{tabular}
\vspace{0.15cm}
\caption{Results of listening test 1. It presents the Mean Opinion Scores (MOS) along with a 95\% confidence interval.}
\label{tab:tab3}

\end{table}
\vspace{-2em}
After proving the validity of our dataset, then we select five different songs corresponding to the model rating of five levels and ask volunteers to subjectively rate them, and this will determine whether our aesthetic model can have good musical aesthetic differentiation ability, shown as Table\ref{tab:tab4}:

\begin{table}[htbp]
\centering
\small
\begin{tabular}{@{}ccccc@{}}
\toprule
\textbf{Low} & \textbf{Medium Low} & \textbf{Medium} & \textbf{Medium High} & \textbf{High} \\ \midrule
1.46 ± 0.17 & 2.65 ± 0.32     & 3.24 ± 0.27 & 3.82 ± 0.31      & 4.32 ± 0.26  \\ \bottomrule
\end{tabular}
\vspace{0.15cm}
\caption{Results of listening test 2. It presents the Mean Opinion Scores (MOS) along with a 95\% confidence interval.}
\label{tab:tab4}
\end{table}
\vspace{-2em}
The subjective experimental results demonstrate that our aesthetic model is effective and can measure the beauty of music. As for the recommendation algorithm, we also briefly ask volunteers to conduct rating experiments, and 93.3\% of volunteers believe that our recommendation algorithm can recommend more aesthetically pleasing music, proving its effectiveness.

\subsection{Ablation Study}

We conduct ablation study to remove harmony, symmetry, chaos, and redundancy to train four different models. We compare them with our full model. The following Table\ref{tab:tab5} shows the results:

\begin{table}[htbp]
\centering
\tabcolsep=0.65cm
\begin{tabular}{@{}ll@{}}
\toprule
\textbf{Our Full Model} & 0.674 \\ \midrule
  \ \ \ \ -w/o harmony    & 1.028                      \\
  \ \ \ \ -w/o symmetry   & 0.723                      \\
  \ \ \ \ -w/o chaos      & 0.785                      \\
  \ \ \ \ -w/o redundancy & 0.836                      \\ \bottomrule
\end{tabular}
\vspace{0.15cm}
\caption{Comparisons of the final MSE between the ablation models and the ground truths, in log scales.}
\label{tab:tab5}
\end{table}
\vspace{-2em}

Experiments have shown that our full model has the best performance, with harmony being the most important aesthetic feature. Regarding the performance ablation of the model, after removing aes features, there is no significant performance decline .
\section{Conclusion}

In this article, we propose an O/C based music aesthetic evaluation model and propose some music features. In addition, we also integrate our music aesthetic features for music aesthetic recommendation task. Nevertheless, there are still shortcomings in this work, such as the lack of consideration of "innovation". We hope that this article may help improve the AIGC music generation task.

\begin{acks}
    We thank the ACs, reviewers and the AI Music Group of BIGAI. This work is partially supported by the Natural Science Foundation of China (62072014\&62106118), the Fundamental Research Funds for the Central Universities (3282023014), the Project of Philosophy and Social Science Research, Ministry of Education of China (20YJC760115), and the Science and Technology Project of the State Archives Administrator (2022-X-069).
\end{acks}

\appendix
\section{Pesudo Features}

\begin{figure}[htbp]
    \centering
      \includegraphics[width=0.5\textwidth]{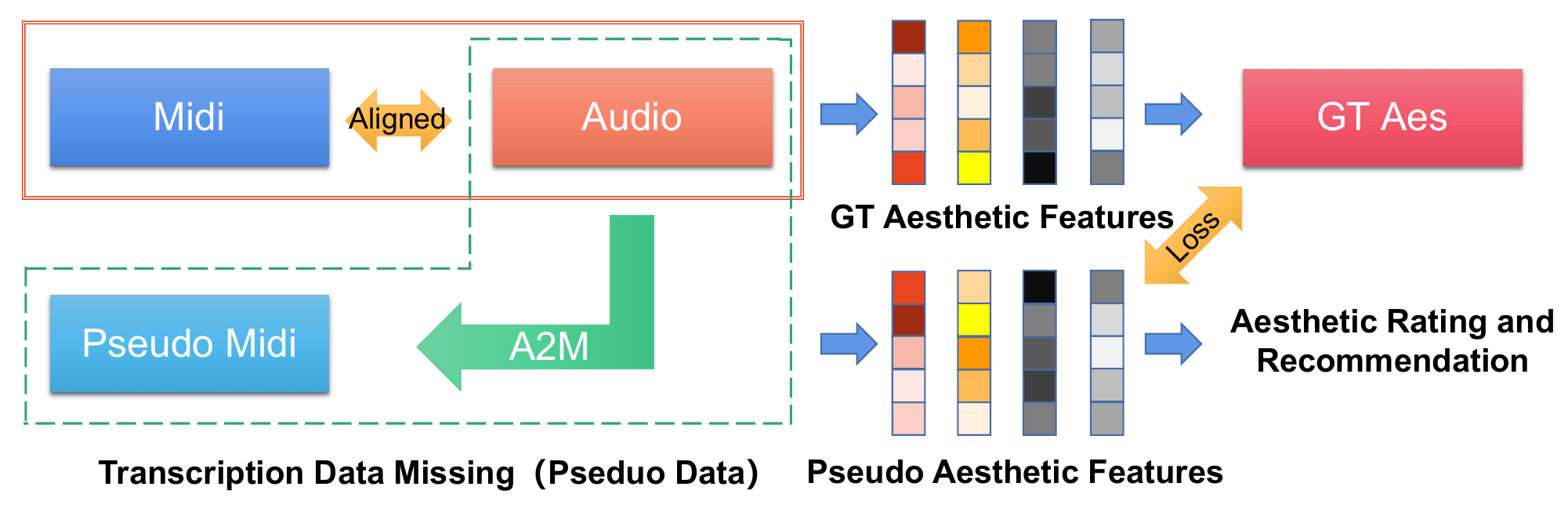}
      \caption{From aesthetic features to pesudo features, the actual features used for aesthetic scoring and recommendation are actually pesudo features.}
      \label{fig:a2m}
      \vspace{-0.5em}
    \end{figure}

Due to the fact that our music basic features include both midi and audio features, the data to be evaluated in the model is often audio data, and the data to be evaluated cannot to be aligned. So, in order to solve the problem of only audio input data, we have designed a method to create pesudo features to replace aesthetic features to complete the following aesthetic calculation tasks (In Fig\ref{fig:a2m}).

Firstly, we train the model using aligned midi and audio data, and calculate the values of four aesthetic features using the algorithms in the aesthetic model. 

Next, we use the a2m algorithm to transcribe audio data with missing aligned midi data, resulting in pesudo midi data.

Then, we treat pesudo midi and audio data as aligned data to extract pesudo features and evaluate a pesudo rating.

Finally, we use 4 aesthetic features and 4 pesudo features as loss, while using both aesthetic and pesudo ratings as loss to make the value of pesudo data as close as possible to the value of GT data.

\section{Extra Experimental Results}

Below are experimental data and results of the pesudo features task:
\vspace{-0.5cm}
\begin{table}[htbp]
\renewcommand\arraystretch{1.2}
\begin{tabular}{@{}cccc@{}}
\toprule

\textbf{Metric}     & \textbf{Aligned M \& A} & \textbf{Pesudo M \& A} & \textbf{A2M M \& A} \\ \midrule
Accuracy   & 92.3\%                & 91.7\%               & 84.2\%            \\
F1-Measure & 90.5\%                & 90.1\%               & 79.8\%            \\ \bottomrule
\end{tabular}
\vspace{0.2cm}
\captionsetup{font={bf,stretch=1.15}}
\caption{The table compares the performance of three types of data in aesthetic rating tasks. From left to right, they are: Aligned Midi and Audio data, Pesudo Midi and Audio data, and A2M (method in previous) Midi and Audio data.}
\label{tab:tab6}
\end{table}
\vspace{-1.1cm}
\begin{table}[htbp]
\small
\renewcommand\arraystretch{1.3}
\begin{tabular}{@{}cccccc@{}}
\toprule
\textbf{Dataset} & \textbf{HR@1} & \textbf{HR@5} & \textbf{HR@10} & \textbf{NDCG@5} & \textbf{NDCG@10} \\ \midrule
MSD              & 0.0931        & 0.2188        & 0.3076         & 0.1633          & 0.1892           \\
Last.fm-1k       & 0.1017        & 0.2699        & 0.4034         & 0.1911          & 0.2197           \\
Last.fm-360k     & 0.2928        & 0.5925        & 0.6980         & 0.4351          & 0.4988           \\ \bottomrule
\end{tabular}
\vspace{0.2cm}
\captionsetup{font={bf,stretch=1.15}}
\caption{The table data shows the performance of using CL4SRec as the backbone network to fuse pesudo aesthetic features in recommendation tasks.}
\label{tab:tab7}
\end{table}

\bibliographystyle{ACM-Reference-Format}
\bibliography{sample-base}


\begin{thebibliography}{72}


\ifx \showCODEN    \undefined \def \showCODEN     #1{\unskip}     \fi
\ifx \showDOI      \undefined \def \showDOI       #1{#1}\fi
\ifx \showISBNx    \undefined \def \showISBNx     #1{\unskip}     \fi
\ifx \showISBNxiii \undefined \def \showISBNxiii  #1{\unskip}     \fi
\ifx \showISSN     \undefined \def \showISSN      #1{\unskip}     \fi
\ifx \showLCCN     \undefined \def \showLCCN      #1{\unskip}     \fi
\ifx \shownote     \undefined \def \shownote      #1{#1}          \fi
\ifx \showarticletitle \undefined \def \showarticletitle #1{#1}   \fi
\ifx \showURL      \undefined \def \showURL       {\relax}        \fi
\providecommand\bibfield[2]{#2}
\providecommand\bibinfo[2]{#2}
\providecommand\natexlab[1]{#1}
\providecommand\showeprint[2][]{arXiv:#2}

\bibitem[al~Rifaie et~al\mbox{.}(2017)]%
        {al2017symmetry}
\bibfield{author}{\bibinfo{person}{Mohammad~Majid al Rifaie},
  \bibinfo{person}{Anna Ursyn}, \bibinfo{person}{Robert Zimmer}, {and}
  \bibinfo{person}{Mohammad Ali~Javaheri Javid}.}
  \bibinfo{year}{2017}\natexlab{}.
\newblock \showarticletitle{On symmetry, aesthetics and quantifying symmetrical
  complexity}. In \bibinfo{booktitle}{\emph{Computational Intelligence in
  Music, Sound, Art and Design: 6th International Conference, EvoMUSART 2017,
  Amsterdam, The Netherlands, April 19--21, 2017, Proceedings 6}}. Springer,
  \bibinfo{pages}{17--32}.
\newblock


\bibitem[Baltrunas et~al\mbox{.}(2011)]%
        {baltrunas2011incarmusic}
\bibfield{author}{\bibinfo{person}{Linas Baltrunas}, \bibinfo{person}{Marius
  Kaminskas}, \bibinfo{person}{Bernd Ludwig}, \bibinfo{person}{Omar Moling},
  \bibinfo{person}{Francesco Ricci}, \bibinfo{person}{Aykan Aydin},
  \bibinfo{person}{Karl-Heinz L{\"u}ke}, {and} \bibinfo{person}{Roland
  Schwaiger}.} \bibinfo{year}{2011}\natexlab{}.
\newblock \showarticletitle{Incarmusic: Context-aware music recommendations in
  a car}. In \bibinfo{booktitle}{\emph{E-Commerce and Web Technologies: 12th
  International Conference, EC-Web 2011, Toulouse, France, August 30-September
  1, 2011. Proceedings 12}}. Springer, \bibinfo{pages}{89--100}.
\newblock


\bibitem[Bense(1960)]%
        {bense1960programmierung}
\bibfield{author}{\bibinfo{person}{Max Bense}.}
  \bibinfo{year}{1960}\natexlab{}.
\newblock \showarticletitle{Programmierung des Sch{\"o}nen Allgemeine
  Texttheorie Und Text{\"a}sthetik}.
\newblock  (\bibinfo{year}{1960}).
\newblock


\bibitem[Bertin-Mahieux et~al\mbox{.}(2011)]%
        {Bertin-Mahieux2011}
\bibfield{author}{\bibinfo{person}{Thierry Bertin-Mahieux},
  \bibinfo{person}{Daniel~P.W. Ellis}, \bibinfo{person}{Brian Whitman}, {and}
  \bibinfo{person}{Paul Lamere}.} \bibinfo{year}{2011}\natexlab{}.
\newblock \showarticletitle{The Million Song Dataset}. In
  \bibinfo{booktitle}{\emph{{Proceedings of the 12th International Conference
  on Music Information Retrieval ({ISMIR} 2011)}}}.
\newblock


\bibitem[Birkhoff(2013)]%
        {birkhoff2013aesthetic}
\bibfield{author}{\bibinfo{person}{George~David Birkhoff}.}
  \bibinfo{year}{2013}\natexlab{}.
\newblock \showarticletitle{Aesthetic measure}.
\newblock In \bibinfo{booktitle}{\emph{Aesthetic Measure}}.
  \bibinfo{publisher}{Harvard University Press}.
\newblock


\bibitem[Bittner et~al\mbox{.}(2014)]%
        {Bittner2014}
\bibfield{author}{\bibinfo{person}{Rachel~M. Bittner}, \bibinfo{person}{Justin
  Salamon}, \bibinfo{person}{Mike Tierney}, \bibinfo{person}{Matthias Mauch},
  \bibinfo{person}{Chris Cannam}, {and} \bibinfo{person}{Juan~Pablo Bello}.}
  \bibinfo{year}{2014}\natexlab{}.
\newblock \showarticletitle{MedleyDB: A Multitrack Dataset for
  Annotation-Intensive MIR Research}.
\newblock \bibinfo{journal}{\emph{International Society for Music Information
  Retrieval Conference}} (\bibinfo{year}{2014}).
\newblock
\urldef\tempurl%
\url{https://www.upf.edu/web/mtg/multitrack-databases}
\showURL{%
\tempurl}


\bibitem[Bonada et~al\mbox{.}(2016)]%
        {bonada2016expressive}
\bibfield{author}{\bibinfo{person}{Jordi Bonada}, \bibinfo{person}{Mart{\'\i}
  Umbert~Morist}, {and} \bibinfo{person}{Merlijn Blaauw}.}
  \bibinfo{year}{2016}\natexlab{}.
\newblock \showarticletitle{Expressive singing synthesis based on unit
  selection for the singing synthesis challenge 2016}.
\newblock \bibinfo{journal}{\emph{Morgan N, editor. Interspeech 2016; 2016 Sep
  8-12; San Francisco, CA.[place unknown]: ISCA; 2016. p. 1230-4.}}
  (\bibinfo{year}{2016}).
\newblock


\bibitem[Brunner et~al\mbox{.}(2018)]%
        {brunner2018midi}
\bibfield{author}{\bibinfo{person}{Gino Brunner}, \bibinfo{person}{Andres
  Konrad}, \bibinfo{person}{Yuyi Wang}, {and} \bibinfo{person}{Roger
  Wattenhofer}.} \bibinfo{year}{2018}\natexlab{}.
\newblock \showarticletitle{MIDI-VAE: Modeling dynamics and instrumentation of
  music with applications to style transfer}.
\newblock \bibinfo{journal}{\emph{arXiv preprint arXiv:1809.07600}}
  (\bibinfo{year}{2018}).
\newblock


\bibitem[Cantador et~al\mbox{.}(2011)]%
        {Cantador2011}
\bibfield{author}{\bibinfo{person}{Iván Cantador}, \bibinfo{person}{Peter
  Brusilovsky}, {and} \bibinfo{person}{Tsvi Kuflik}.}
  \bibinfo{year}{2011}\natexlab{}.
\newblock \showarticletitle{2nd Workshop on Information Heterogeneity and
  Fusion in Recommender Systems (HetRec 2011)}.
\newblock \bibinfo{journal}{\emph{Proceedings of the 5th ACM Conference on
  Recommender Systems}} (\bibinfo{year}{2011}), \bibinfo{pages}{387--388}.
\newblock
\urldef\tempurl%
\url{https://doi.org/10.1145/2043932.2044017}
\showDOI{\tempurl}


\bibitem[Celma(2008)]%
        {Celma2008}
\bibfield{author}{\bibinfo{person}{Òscar Celma}.}
  \bibinfo{year}{2008}\natexlab{}.
\newblock \showarticletitle{Music Recommendation and Discovery in the Long
  Tail}. In \bibinfo{booktitle}{\emph{Proceedings of the 1st ACM International
  Conference on Recommender Systems}}. \bibinfo{pages}{5--12}.
\newblock
\urldef\tempurl%
\url{https://doi.org/10.1145/1454008.1454013}
\showDOI{\tempurl}


\bibitem[C{\'\i}fka et~al\mbox{.}(2019)]%
        {cifka2019supervised}
\bibfield{author}{\bibinfo{person}{Ond{\v{r}}ej C{\'\i}fka},
  \bibinfo{person}{Umut {\c{S}}im{\c{s}}ekli}, {and} \bibinfo{person}{Ga{\"e}l
  Richard}.} \bibinfo{year}{2019}\natexlab{}.
\newblock \showarticletitle{Supervised symbolic music style translation using
  synthetic data}.
\newblock \bibinfo{journal}{\emph{arXiv preprint arXiv:1907.02265}}
  (\bibinfo{year}{2019}).
\newblock


\bibitem[Clemente et~al\mbox{.}(2022)]%
        {clemente2022musical}
\bibfield{author}{\bibinfo{person}{Ana Clemente}, \bibinfo{person}{Marcus~T
  Pearce}, {and} \bibinfo{person}{Marcos Nadal}.}
  \bibinfo{year}{2022}\natexlab{}.
\newblock \showarticletitle{Musical aesthetic sensitivity.}
\newblock \bibinfo{journal}{\emph{Psychology of Aesthetics, Creativity, and the
  Arts}} \bibinfo{volume}{16}, \bibinfo{number}{1} (\bibinfo{year}{2022}),
  \bibinfo{pages}{58}.
\newblock


\bibitem[Deng et~al\mbox{.}(2017)]%
        {deng2017image}
\bibfield{author}{\bibinfo{person}{Yubin Deng}, \bibinfo{person}{Chen~Change
  Loy}, {and} \bibinfo{person}{Xiaoou Tang}.} \bibinfo{year}{2017}\natexlab{}.
\newblock \showarticletitle{Image aesthetic assessment: An experimental
  survey}.
\newblock \bibinfo{journal}{\emph{IEEE Signal Processing Magazine}}
  \bibinfo{volume}{34}, \bibinfo{number}{4} (\bibinfo{year}{2017}),
  \bibinfo{pages}{80--106}.
\newblock


\bibitem[Dieleman et~al\mbox{.}(2018)]%
        {dieleman2018challenge}
\bibfield{author}{\bibinfo{person}{Sander Dieleman}, \bibinfo{person}{Aaron
  van~den Oord}, {and} \bibinfo{person}{Karen Simonyan}.}
  \bibinfo{year}{2018}\natexlab{}.
\newblock \showarticletitle{The challenge of realistic music generation:
  modelling raw audio at scale}.
\newblock \bibinfo{journal}{\emph{Advances in Neural Information Processing
  Systems}}  \bibinfo{volume}{31} (\bibinfo{year}{2018}).
\newblock


\bibitem[Dong et~al\mbox{.}(2018)]%
        {dong2018musegan}
\bibfield{author}{\bibinfo{person}{Hao-Wen Dong}, \bibinfo{person}{Wen-Yi
  Hsiao}, \bibinfo{person}{Li-Chia Yang}, {and} \bibinfo{person}{Yi-Hsuan
  Yang}.} \bibinfo{year}{2018}\natexlab{}.
\newblock \showarticletitle{Musegan: Multi-track sequential generative
  adversarial networks for symbolic music generation and accompaniment}. In
  \bibinfo{booktitle}{\emph{Proceedings of the AAAI Conference on Artificial
  Intelligence}}, Vol.~\bibinfo{volume}{32}.
\newblock


\bibitem[Dong et~al\mbox{.}(2022)]%
        {dong2022deep}
\bibfield{author}{\bibinfo{person}{Hao-Wen Dong}, \bibinfo{person}{Cong Zhou},
  \bibinfo{person}{Taylor Berg-Kirkpatrick}, {and} \bibinfo{person}{Julian
  McAuley}.} \bibinfo{year}{2022}\natexlab{}.
\newblock \showarticletitle{Deep performer: Score-to-audio music performance
  synthesis}. In \bibinfo{booktitle}{\emph{ICASSP 2022-2022 IEEE International
  Conference on Acoustics, Speech and Signal Processing (ICASSP)}}. IEEE,
  \bibinfo{pages}{951--955}.
\newblock


\bibitem[Dubnov et~al\mbox{.}(2011)]%
        {dubnov2011audio}
\bibfield{author}{\bibinfo{person}{Shlomo Dubnov}, \bibinfo{person}{G{\'e}rard
  Assayag}, {and} \bibinfo{person}{Arshia Cont}.}
  \bibinfo{year}{2011}\natexlab{}.
\newblock \showarticletitle{Audio oracle analysis of musical information rate}.
  In \bibinfo{booktitle}{\emph{2011 ieee fifth international conference on
  semantic computing}}. IEEE, \bibinfo{pages}{567--571}.
\newblock


\bibitem[Engel et~al\mbox{.}(2019)]%
        {engel2019gansynth}
\bibfield{author}{\bibinfo{person}{Jesse Engel}, \bibinfo{person}{Kumar~Krishna
  Agrawal}, \bibinfo{person}{Shuo Chen}, \bibinfo{person}{Ishaan Gulrajani},
  \bibinfo{person}{Chris Donahue}, {and} \bibinfo{person}{Adam Roberts}.}
  \bibinfo{year}{2019}\natexlab{}.
\newblock \showarticletitle{Gansynth: Adversarial neural audio synthesis}.
\newblock \bibinfo{journal}{\emph{arXiv preprint arXiv:1902.08710}}
  (\bibinfo{year}{2019}).
\newblock


\bibitem[Engel et~al\mbox{.}(2017)]%
        {engel2017neural}
\bibfield{author}{\bibinfo{person}{Jesse Engel}, \bibinfo{person}{Cinjon
  Resnick}, \bibinfo{person}{Adam Roberts}, \bibinfo{person}{Sander Dieleman},
  \bibinfo{person}{Mohammad Norouzi}, \bibinfo{person}{Douglas Eck}, {and}
  \bibinfo{person}{Karen Simonyan}.} \bibinfo{year}{2017}\natexlab{}.
\newblock \showarticletitle{Neural audio synthesis of musical notes with
  wavenet autoencoders}. In \bibinfo{booktitle}{\emph{International Conference
  on Machine Learning}}. PMLR, \bibinfo{pages}{1068--1077}.
\newblock


\bibitem[Galanter(2012)]%
        {galanter2012computational}
\bibfield{author}{\bibinfo{person}{Philip Galanter}.}
  \bibinfo{year}{2012}\natexlab{}.
\newblock \showarticletitle{Computational aesthetic evaluation: steps towards
  machine creativity}.
\newblock In \bibinfo{booktitle}{\emph{ACM SIGGRAPH 2012 Courses}}.
  \bibinfo{pages}{1--162}.
\newblock


\bibitem[Galanter(2013)]%
        {galanter2013computational}
\bibfield{author}{\bibinfo{person}{Philip Galanter}.}
  \bibinfo{year}{2013}\natexlab{}.
\newblock \showarticletitle{Computational aesthetic evaluation: Automated
  fitness functions for evolutionary art, design, and music}. In
  \bibinfo{booktitle}{\emph{Proceedings of the 15th annual conference companion
  on Genetic and evolutionary computation}}. \bibinfo{pages}{1005--1038}.
\newblock


\bibitem[Hadjeres et~al\mbox{.}(2017)]%
        {hadjeres2017deepbach}
\bibfield{author}{\bibinfo{person}{Ga{\"e}tan Hadjeres},
  \bibinfo{person}{Fran{\c{c}}ois Pachet}, {and} \bibinfo{person}{Frank
  Nielsen}.} \bibinfo{year}{2017}\natexlab{}.
\newblock \showarticletitle{Deepbach: a steerable model for bach chorales
  generation}. In \bibinfo{booktitle}{\emph{International Conference on Machine
  Learning}}. PMLR, \bibinfo{pages}{1362--1371}.
\newblock


\bibitem[Haque et~al\mbox{.}(2018)]%
        {haque2018conditional}
\bibfield{author}{\bibinfo{person}{Albert Haque}, \bibinfo{person}{Michelle
  Guo}, {and} \bibinfo{person}{Prateek Verma}.}
  \bibinfo{year}{2018}\natexlab{}.
\newblock \showarticletitle{Conditional end-to-end audio transforms}.
\newblock \bibinfo{journal}{\emph{arXiv preprint arXiv:1804.00047}}
  (\bibinfo{year}{2018}).
\newblock


\bibitem[Hawthorne et~al\mbox{.}(2017)]%
        {hawthorne2017onsets}
\bibfield{author}{\bibinfo{person}{Curtis Hawthorne}, \bibinfo{person}{Erich
  Elsen}, \bibinfo{person}{Jialin Song}, \bibinfo{person}{Adam Roberts},
  \bibinfo{person}{Ian Simon}, \bibinfo{person}{Colin Raffel},
  \bibinfo{person}{Jesse Engel}, \bibinfo{person}{Sageev Oore}, {and}
  \bibinfo{person}{Douglas Eck}.} \bibinfo{year}{2017}\natexlab{}.
\newblock \showarticletitle{Onsets and frames: Dual-objective piano
  transcription}.
\newblock \bibinfo{journal}{\emph{arXiv preprint arXiv:1710.11153}}
  (\bibinfo{year}{2017}).
\newblock


\bibitem[Hidasi et~al\mbox{.}(2015)]%
        {hidasi2015session}
\bibfield{author}{\bibinfo{person}{Bal{\'a}zs Hidasi},
  \bibinfo{person}{Alexandros Karatzoglou}, \bibinfo{person}{Linas Baltrunas},
  {and} \bibinfo{person}{Domonkos Tikk}.} \bibinfo{year}{2015}\natexlab{}.
\newblock \showarticletitle{Session-based recommendations with recurrent neural
  networks}.
\newblock \bibinfo{journal}{\emph{arXiv preprint arXiv:1511.06939}}
  (\bibinfo{year}{2015}).
\newblock


\bibitem[Huang et~al\mbox{.}(2018)]%
        {huang2018music}
\bibfield{author}{\bibinfo{person}{Cheng-Zhi~Anna Huang},
  \bibinfo{person}{Ashish Vaswani}, \bibinfo{person}{Jakob Uszkoreit},
  \bibinfo{person}{Noam Shazeer}, \bibinfo{person}{Ian Simon},
  \bibinfo{person}{Curtis Hawthorne}, \bibinfo{person}{Andrew~M Dai},
  \bibinfo{person}{Matthew~D Hoffman}, \bibinfo{person}{Monica Dinculescu},
  {and} \bibinfo{person}{Douglas Eck}.} \bibinfo{year}{2018}\natexlab{}.
\newblock \showarticletitle{Music transformer}.
\newblock \bibinfo{journal}{\emph{arXiv preprint arXiv:1809.04281}}
  (\bibinfo{year}{2018}).
\newblock


\bibitem[Huang et~al\mbox{.}({[n.\,d.]})]%
        {huangmusic}
\bibfield{author}{\bibinfo{person}{Cheng-Zhi~Anna Huang},
  \bibinfo{person}{Ashish Vaswani}, \bibinfo{person}{Jakob Uszkoreit},
  \bibinfo{person}{Ian Simon}, \bibinfo{person}{Curtis Hawthorne},
  \bibinfo{person}{Noam Shazeer}, \bibinfo{person}{Andrew~M Dai},
  \bibinfo{person}{Matthew~D Hoffman}, \bibinfo{person}{Monica Dinculescu},
  {and} \bibinfo{person}{Douglas Eck}.} \bibinfo{year}{[n.\,d.]}\natexlab{}.
\newblock \showarticletitle{Music Transformer: Generating Music with Long-Term
  Structure}. In \bibinfo{booktitle}{\emph{International Conference on Learning
  Representations}}.
\newblock


\bibitem[Huang et~al\mbox{.}(2022)]%
        {huang2022singgan}
\bibfield{author}{\bibinfo{person}{Rongjie Huang}, \bibinfo{person}{Chenye
  Cui}, \bibinfo{person}{Feiyang Chen}, \bibinfo{person}{Yi Ren},
  \bibinfo{person}{Jinglin Liu}, \bibinfo{person}{Zhou Zhao},
  \bibinfo{person}{Baoxing Huai}, {and} \bibinfo{person}{Zhefeng Wang}.}
  \bibinfo{year}{2022}\natexlab{}.
\newblock \showarticletitle{Singgan: Generative adversarial network for
  high-fidelity singing voice generation}. In
  \bibinfo{booktitle}{\emph{Proceedings of the 30th ACM International
  Conference on Multimedia}}. \bibinfo{pages}{2525--2535}.
\newblock


\bibitem[Jeong et~al\mbox{.}(2019)]%
        {jeong2019virtuosonet}
\bibfield{author}{\bibinfo{person}{Dasaem Jeong}, \bibinfo{person}{Taegyun
  Kwon}, \bibinfo{person}{Yoojin Kim}, \bibinfo{person}{Kyogu Lee}, {and}
  \bibinfo{person}{Juhan Nam}.} \bibinfo{year}{2019}\natexlab{}.
\newblock \showarticletitle{VirtuosoNet: A Hierarchical RNN-based System for
  Modeling Expressive Piano Performance.}. In
  \bibinfo{booktitle}{\emph{ISMIR}}. \bibinfo{pages}{908--915}.
\newblock


\bibitem[Ji et~al\mbox{.}(2020)]%
        {ji2020comprehensive}
\bibfield{author}{\bibinfo{person}{Shulei Ji}, \bibinfo{person}{Jing Luo},
  {and} \bibinfo{person}{Xinyu Yang}.} \bibinfo{year}{2020}\natexlab{}.
\newblock \showarticletitle{A comprehensive survey on deep music generation:
  Multi-level representations, algorithms, evaluations, and future directions}.
\newblock \bibinfo{journal}{\emph{arXiv preprint arXiv:2011.06801}}
  (\bibinfo{year}{2020}).
\newblock


\bibitem[Jiang et~al\mbox{.}(2017)]%
        {jiang2017play}
\bibfield{author}{\bibinfo{person}{Miao Jiang}, \bibinfo{person}{Ziyi Yang},
  {and} \bibinfo{person}{Chen Zhao}.} \bibinfo{year}{2017}\natexlab{}.
\newblock \showarticletitle{What to play next? A RNN-based music recommendation
  system}. In \bibinfo{booktitle}{\emph{2017 51st Asilomar Conference on
  Signals, Systems, and Computers}}. IEEE, \bibinfo{pages}{356--358}.
\newblock


\bibitem[Jin et~al\mbox{.}(2019)]%
        {jin2019aesthetic}
\bibfield{author}{\bibinfo{person}{Xin Jin}, \bibinfo{person}{Le Wu},
  \bibinfo{person}{Geng Zhao}, \bibinfo{person}{Xiaodong Li},
  \bibinfo{person}{Xiaokun Zhang}, \bibinfo{person}{Shiming Ge},
  \bibinfo{person}{Dongqing Zou}, \bibinfo{person}{Bin Zhou}, {and}
  \bibinfo{person}{Xinghui Zhou}.} \bibinfo{year}{2019}\natexlab{}.
\newblock \showarticletitle{Aesthetic attributes assessment of images}. In
  \bibinfo{booktitle}{\emph{Proceedings of the 27th ACM international
  conference on multimedia}}. \bibinfo{pages}{311--319}.
\newblock


\bibitem[Jin et~al\mbox{.}(2023a)]%
        {jin2023order}
\bibfield{author}{\bibinfo{person}{Xin Jin}, \bibinfo{person}{Wu Zhou},
  \bibinfo{person}{Jinyu Wang}, \bibinfo{person}{Duo Xu},
  \bibinfo{person}{Yiqing Rong}, {and} \bibinfo{person}{Shuai Cui}.}
  \bibinfo{year}{2023}\natexlab{a}.
\newblock \showarticletitle{An Order-Complexity Model for Aesthetic Quality
  Assessment of Symbolic Homophony Music Scores}.
\newblock \bibinfo{journal}{\emph{arXiv preprint arXiv:2301.05908}}
  (\bibinfo{year}{2023}).
\newblock


\bibitem[Jin et~al\mbox{.}(2023b)]%
        {jin2023order2}
\bibfield{author}{\bibinfo{person}{Xin Jin}, \bibinfo{person}{Wu Zhou},
  \bibinfo{person}{Jinyu Wang}, \bibinfo{person}{Duo Xu},
  \bibinfo{person}{Yiqing Rong}, {and} \bibinfo{person}{Jialin Sun}.}
  \bibinfo{year}{2023}\natexlab{b}.
\newblock \showarticletitle{An Order-Complexity Model for Aesthetic Quality
  Assessment of Homophony Music Performance}.
\newblock \bibinfo{journal}{\emph{arXiv preprint arXiv:2304.11521}}
  (\bibinfo{year}{2023}).
\newblock


\bibitem[Kang and McAuley(2018)]%
        {kang2018self}
\bibfield{author}{\bibinfo{person}{Wang-Cheng Kang} {and}
  \bibinfo{person}{Julian McAuley}.} \bibinfo{year}{2018}\natexlab{}.
\newblock \showarticletitle{Self-attentive sequential recommendation}. In
  \bibinfo{booktitle}{\emph{2018 IEEE international conference on data mining
  (ICDM)}}. IEEE, \bibinfo{pages}{197--206}.
\newblock


\bibitem[Kingma and Ba(2014)]%
        {kingma2014adam}
\bibfield{author}{\bibinfo{person}{Diederik~P Kingma} {and}
  \bibinfo{person}{Jimmy Ba}.} \bibinfo{year}{2014}\natexlab{}.
\newblock \showarticletitle{Adam: A method for stochastic optimization}.
\newblock \bibinfo{journal}{\emph{arXiv preprint arXiv:1412.6980}}
  (\bibinfo{year}{2014}).
\newblock


\bibitem[Kong et~al\mbox{.}(2016)]%
        {kong2016photo}
\bibfield{author}{\bibinfo{person}{Shu Kong}, \bibinfo{person}{Xiaohui Shen},
  \bibinfo{person}{Zhe Lin}, \bibinfo{person}{Radomir Mech}, {and}
  \bibinfo{person}{Charless Fowlkes}.} \bibinfo{year}{2016}\natexlab{}.
\newblock \showarticletitle{Photo aesthetics ranking network with attributes
  and content adaptation}. In \bibinfo{booktitle}{\emph{Computer Vision--ECCV
  2016: 14th European Conference, Amsterdam, The Netherlands, October 11--14,
  2016, Proceedings, Part I 14}}. Springer, \bibinfo{pages}{662--679}.
\newblock


\bibitem[Kulkarni and Rodd(2020)]%
        {kulkarni2020context}
\bibfield{author}{\bibinfo{person}{Saurabh Kulkarni} {and}
  \bibinfo{person}{Sunil~F Rodd}.} \bibinfo{year}{2020}\natexlab{}.
\newblock \showarticletitle{Context Aware Recommendation Systems: A review of
  the state of the art techniques}.
\newblock \bibinfo{journal}{\emph{Computer Science Review}}
  \bibinfo{volume}{37} (\bibinfo{year}{2020}), \bibinfo{pages}{100255}.
\newblock


\bibitem[Lerdahl and Jackendoff(1996)]%
        {lerdahl1996generative}
\bibfield{author}{\bibinfo{person}{Fred Lerdahl} {and} \bibinfo{person}{Ray~S
  Jackendoff}.} \bibinfo{year}{1996}\natexlab{}.
\newblock \bibinfo{booktitle}{\emph{A Generative Theory of Tonal Music,
  reissue, with a new preface}}.
\newblock \bibinfo{publisher}{MIT press}.
\newblock


\bibitem[Lorand(2002)]%
        {lorand2002aesthetic}
\bibfield{author}{\bibinfo{person}{Ruth Lorand}.}
  \bibinfo{year}{2002}\natexlab{}.
\newblock \bibinfo{booktitle}{\emph{Aesthetic order: a philosophy of order,
  beauty and art}}.
\newblock \bibinfo{publisher}{Routledge}.
\newblock


\bibitem[McFee et~al\mbox{.}(2015)]%
        {mcfee2015librosa}
\bibfield{author}{\bibinfo{person}{Brian McFee}, \bibinfo{person}{Colin
  Raffel}, \bibinfo{person}{Dawen Liang}, \bibinfo{person}{Daniel~P Ellis},
  \bibinfo{person}{Matt McVicar}, \bibinfo{person}{Eric Battenberg}, {and}
  \bibinfo{person}{Oriol Nieto}.} \bibinfo{year}{2015}\natexlab{}.
\newblock \showarticletitle{librosa: Audio and music signal analysis in
  python}. In \bibinfo{booktitle}{\emph{Proceedings of the 14th python in
  science conference}}, Vol.~\bibinfo{volume}{8}. \bibinfo{pages}{18--25}.
\newblock


\bibitem[McKay et~al\mbox{.}(2018)]%
        {mckay2018jsymbolic}
\bibfield{author}{\bibinfo{person}{Cory McKay}, \bibinfo{person}{Julie
  Cumming}, {and} \bibinfo{person}{Ichiro Fujinaga}.}
  \bibinfo{year}{2018}\natexlab{}.
\newblock \showarticletitle{JSYMBOLIC 2.2: Extracting Features from Symbolic
  Music for use in Musicological and MIR Research.}. In
  \bibinfo{booktitle}{\emph{ISMIR}}. \bibinfo{pages}{348--354}.
\newblock


\bibitem[McKay et~al\mbox{.}(2005)]%
        {mckay2005jaudio}
\bibfield{author}{\bibinfo{person}{Cory McKay}, \bibinfo{person}{Ichiro
  Fujinaga}, {and} \bibinfo{person}{Philippe Depalle}.}
  \bibinfo{year}{2005}\natexlab{}.
\newblock \showarticletitle{jAudio: A feature extraction library}. In
  \bibinfo{booktitle}{\emph{Proceedings of the international conference on
  music information retrieval}}, Vol.~\bibinfo{volume}{11}.
  \bibinfo{pages}{600--3}.
\newblock


\bibitem[McLeod and Steedman(2018)]%
        {mcleod2018evaluating}
\bibfield{author}{\bibinfo{person}{Andrew McLeod} {and} \bibinfo{person}{Mark
  Steedman}.} \bibinfo{year}{2018}\natexlab{}.
\newblock \showarticletitle{Evaluating Automatic Polyphonic Music
  Transcription.}. In \bibinfo{booktitle}{\emph{ISMIR}}.
  \bibinfo{pages}{42--49}.
\newblock


\bibitem[Mehri et~al\mbox{.}(2016)]%
        {mehri2016samplernn}
\bibfield{author}{\bibinfo{person}{Soroush Mehri}, \bibinfo{person}{Kundan
  Kumar}, \bibinfo{person}{Ishaan Gulrajani}, \bibinfo{person}{Rithesh Kumar},
  \bibinfo{person}{Shubham Jain}, \bibinfo{person}{Jose Sotelo},
  \bibinfo{person}{Aaron Courville}, {and} \bibinfo{person}{Yoshua Bengio}.}
  \bibinfo{year}{2016}\natexlab{}.
\newblock \showarticletitle{SampleRNN: An unconditional end-to-end neural audio
  generation model}.
\newblock \bibinfo{journal}{\emph{arXiv preprint arXiv:1612.07837}}
  (\bibinfo{year}{2016}).
\newblock


\bibitem[Meyer(1956)]%
        {meyer1956meaning}
\bibfield{author}{\bibinfo{person}{Leonard~B Meyer}.}
  \bibinfo{year}{1956}\natexlab{}.
\newblock \bibinfo{title}{Meaning and emotion in music}.
\newblock
\newblock


\bibitem[M{\"u}ller et~al\mbox{.}(2013)]%
        {MuellerJG13_StructureAnaylsis_IEEE-TASLP}
\bibfield{author}{\bibinfo{person}{Meinard M{\"u}ller}, \bibinfo{person}{Nanzhu
  Jiang}, {and} \bibinfo{person}{Peter Grosche}.}
  \bibinfo{year}{2013}\natexlab{}.
\newblock \showarticletitle{A Robust Fitness Measure for Capturing Repetitions
  in Music Recordings With Applications to Audio Thumbnailing}.
\newblock \bibinfo{journal}{\emph{IEEE Transactions on Audio, Speech, and
  Language Processing}} \bibinfo{volume}{21}, \bibinfo{number}{3}
  (\bibinfo{year}{2013}), \bibinfo{pages}{531--543}.
\newblock


\bibitem[Murray et~al\mbox{.}(2012)]%
        {murray2012ava}
\bibfield{author}{\bibinfo{person}{Naila Murray}, \bibinfo{person}{Luca
  Marchesotti}, {and} \bibinfo{person}{Florent Perronnin}.}
  \bibinfo{year}{2012}\natexlab{}.
\newblock \showarticletitle{AVA: A large-scale database for aesthetic visual
  analysis}. In \bibinfo{booktitle}{\emph{2012 IEEE conference on computer
  vision and pattern recognition}}. IEEE, \bibinfo{pages}{2408--2415}.
\newblock


\bibitem[Navarro-C{\'a}ceres et~al\mbox{.}(2020)]%
        {navarro2020computational}
\bibfield{author}{\bibinfo{person}{Mar{\'\i}a Navarro-C{\'a}ceres},
  \bibinfo{person}{Marcelo Caetano}, \bibinfo{person}{Gilberto Bernardes},
  \bibinfo{person}{Mercedes S{\'a}nchez-Barba}, {and} \bibinfo{person}{Javier
  Merch{\'a}n S{\'a}nchez-Jara}.} \bibinfo{year}{2020}\natexlab{}.
\newblock \showarticletitle{A computational model of tonal tension profile of
  chord progressions in the tonal interval space}.
\newblock \bibinfo{journal}{\emph{Entropy}} \bibinfo{volume}{22},
  \bibinfo{number}{11} (\bibinfo{year}{2020}), \bibinfo{pages}{1291}.
\newblock


\bibitem[Oord et~al\mbox{.}(2016)]%
        {oord2016wavenet}
\bibfield{author}{\bibinfo{person}{Aaron van~den Oord}, \bibinfo{person}{Sander
  Dieleman}, \bibinfo{person}{Heiga Zen}, \bibinfo{person}{Karen Simonyan},
  \bibinfo{person}{Oriol Vinyals}, \bibinfo{person}{Alex Graves},
  \bibinfo{person}{Nal Kalchbrenner}, \bibinfo{person}{Andrew Senior}, {and}
  \bibinfo{person}{Koray Kavukcuoglu}.} \bibinfo{year}{2016}\natexlab{}.
\newblock \showarticletitle{Wavenet: A generative model for raw audio}.
\newblock \bibinfo{journal}{\emph{arXiv preprint arXiv:1609.03499}}
  (\bibinfo{year}{2016}).
\newblock


\bibitem[Oore et~al\mbox{.}(2020)]%
        {oore2020time}
\bibfield{author}{\bibinfo{person}{Sageev Oore}, \bibinfo{person}{Ian Simon},
  \bibinfo{person}{Sander Dieleman}, \bibinfo{person}{Douglas Eck}, {and}
  \bibinfo{person}{Karen Simonyan}.} \bibinfo{year}{2020}\natexlab{}.
\newblock \showarticletitle{This time with feeling: Learning expressive musical
  performance}.
\newblock \bibinfo{journal}{\emph{Neural Computing and Applications}}
  \bibinfo{volume}{32} (\bibinfo{year}{2020}), \bibinfo{pages}{955--967}.
\newblock


\bibitem[Osborne(1986)]%
        {osborne1986symmetry}
\bibfield{author}{\bibinfo{person}{Harold Osborne}.}
  \bibinfo{year}{1986}\natexlab{}.
\newblock \showarticletitle{Symmetry as an aesthetic factor}.
\newblock \bibinfo{journal}{\emph{Computers \& Mathematics with Applications}}
  \bibinfo{volume}{12}, \bibinfo{number}{1-2} (\bibinfo{year}{1986}),
  \bibinfo{pages}{77--82}.
\newblock


\bibitem[Oura et~al\mbox{.}(2010)]%
        {oura2010recent}
\bibfield{author}{\bibinfo{person}{Keiichiro Oura}, \bibinfo{person}{Ayami
  Mase}, \bibinfo{person}{Tomohiko Yamada}, \bibinfo{person}{Satoru Muto},
  \bibinfo{person}{Yoshihiko Nankaku}, {and} \bibinfo{person}{Keiichi Tokuda}.}
  \bibinfo{year}{2010}\natexlab{}.
\newblock \showarticletitle{Recent development of the HMM-based singing voice
  synthesis system—Sinsy}. In \bibinfo{booktitle}{\emph{Seventh ISCA Workshop
  on Speech Synthesis}}.
\newblock


\bibitem[Raffel et~al\mbox{.}(2016)]%
        {raffel2016learning}
\bibfield{author}{\bibinfo{person}{Colin Raffel}, \bibinfo{person}{Ron Weiss},
  \bibinfo{person}{Johan Pauwels}, \bibinfo{person}{William Colombo},
  \bibinfo{person}{Curtis Hawthorne}, {and} \bibinfo{person}{Douglas Eck}.}
  \bibinfo{year}{2016}\natexlab{}.
\newblock \showarticletitle{Learning-based methods for comparing sequences,
  with applications to audio-to-MIDI alignment and matching}. In
  \bibinfo{booktitle}{\emph{International Conference on Machine Learning}}.
  PMLR, \bibinfo{pages}{1368--1377}.
\newblock


\bibitem[Rendle et~al\mbox{.}(2012)]%
        {rendle2012bpr}
\bibfield{author}{\bibinfo{person}{Steffen Rendle}, \bibinfo{person}{Christoph
  Freudenthaler}, \bibinfo{person}{Zeno Gantner}, {and} \bibinfo{person}{Lars
  Schmidt-Thieme}.} \bibinfo{year}{2012}\natexlab{}.
\newblock \showarticletitle{BPR: Bayesian personalized ranking from implicit
  feedback}.
\newblock \bibinfo{journal}{\emph{arXiv preprint arXiv:1205.2618}}
  (\bibinfo{year}{2012}).
\newblock


\bibitem[Rendle et~al\mbox{.}(2010)]%
        {rendle2010factorizing}
\bibfield{author}{\bibinfo{person}{Steffen Rendle}, \bibinfo{person}{Christoph
  Freudenthaler}, {and} \bibinfo{person}{Lars Schmidt-Thieme}.}
  \bibinfo{year}{2010}\natexlab{}.
\newblock \showarticletitle{Factorizing personalized markov chains for
  next-basket recommendation}. In \bibinfo{booktitle}{\emph{Proceedings of the
  19th international conference on World wide web}}. \bibinfo{pages}{811--820}.
\newblock


\bibitem[Roberts et~al\mbox{.}(2018)]%
        {roberts2018hierarchical}
\bibfield{author}{\bibinfo{person}{Adam Roberts}, \bibinfo{person}{Jesse
  Engel}, \bibinfo{person}{Colin Raffel}, \bibinfo{person}{Curtis Hawthorne},
  {and} \bibinfo{person}{Douglas Eck}.} \bibinfo{year}{2018}\natexlab{}.
\newblock \showarticletitle{A hierarchical latent vector model for learning
  long-term structure in music}. In \bibinfo{booktitle}{\emph{International
  conference on machine learning}}. PMLR, \bibinfo{pages}{4364--4373}.
\newblock


\bibitem[Sachdeva et~al\mbox{.}(2018)]%
        {sachdeva2018attentive}
\bibfield{author}{\bibinfo{person}{Noveen Sachdeva}, \bibinfo{person}{Kartik
  Gupta}, {and} \bibinfo{person}{Vikram Pudi}.}
  \bibinfo{year}{2018}\natexlab{}.
\newblock \showarticletitle{Attentive neural architecture incorporating song
  features for music recommendation}. In \bibinfo{booktitle}{\emph{Proceedings
  of the 12th ACM Conference on Recommender Systems}}.
  \bibinfo{pages}{417--421}.
\newblock


\bibitem[Schedl et~al\mbox{.}(2015)]%
        {schedl2015music}
\bibfield{author}{\bibinfo{person}{Markus Schedl}, \bibinfo{person}{Peter
  Knees}, \bibinfo{person}{Brian McFee}, \bibinfo{person}{Dmitry Bogdanov},
  {and} \bibinfo{person}{Marius Kaminskas}.} \bibinfo{year}{2015}\natexlab{}.
\newblock \showarticletitle{Music recommender systems}.
\newblock \bibinfo{journal}{\emph{Recommender systems handbook}}
  (\bibinfo{year}{2015}), \bibinfo{pages}{453--492}.
\newblock


\bibitem[Shen et~al\mbox{.}(2020)]%
        {shen2020peia}
\bibfield{author}{\bibinfo{person}{Tiancheng Shen}, \bibinfo{person}{Jia Jia},
  \bibinfo{person}{Yan Li}, \bibinfo{person}{Yihui Ma}, \bibinfo{person}{Yaohua
  Bu}, \bibinfo{person}{Hanjie Wang}, \bibinfo{person}{Bo Chen},
  \bibinfo{person}{Tat-Seng Chua}, {and} \bibinfo{person}{Wendy Hall}.}
  \bibinfo{year}{2020}\natexlab{}.
\newblock \showarticletitle{Peia: Personality and emotion integrated attentive
  model for music recommendation on social media platforms}. In
  \bibinfo{booktitle}{\emph{Proceedings of the AAAI conference on artificial
  intelligence}}, Vol.~\bibinfo{volume}{34}. \bibinfo{pages}{206--213}.
\newblock


\bibitem[Shih et~al\mbox{.}(2022)]%
        {shih2022theme}
\bibfield{author}{\bibinfo{person}{Yi-Jen Shih}, \bibinfo{person}{Shih-Lun Wu},
  \bibinfo{person}{Frank Zalkow}, \bibinfo{person}{Meinard Muller}, {and}
  \bibinfo{person}{Yi-Hsuan Yang}.} \bibinfo{year}{2022}\natexlab{}.
\newblock \showarticletitle{Theme transformer: Symbolic music generation with
  theme-conditioned transformer}.
\newblock \bibinfo{journal}{\emph{IEEE Transactions on Multimedia}}
  (\bibinfo{year}{2022}).
\newblock


\bibitem[Sun et~al\mbox{.}(2019)]%
        {sun2019bert4rec}
\bibfield{author}{\bibinfo{person}{Fei Sun}, \bibinfo{person}{Jun Liu},
  \bibinfo{person}{Jian Wu}, \bibinfo{person}{Changhua Pei},
  \bibinfo{person}{Xiao Lin}, \bibinfo{person}{Wenwu Ou}, {and}
  \bibinfo{person}{Peng Jiang}.} \bibinfo{year}{2019}\natexlab{}.
\newblock \showarticletitle{BERT4Rec: Sequential recommendation with
  bidirectional encoder representations from transformer}. In
  \bibinfo{booktitle}{\emph{Proceedings of the 28th ACM international
  conference on information and knowledge management}}.
  \bibinfo{pages}{1441--1450}.
\newblock


\bibitem[Tang and Wang(2018)]%
        {tang2018personalized}
\bibfield{author}{\bibinfo{person}{Jiaxi Tang} {and} \bibinfo{person}{Ke
  Wang}.} \bibinfo{year}{2018}\natexlab{}.
\newblock \showarticletitle{Personalized top-n sequential recommendation via
  convolutional sequence embedding}. In \bibinfo{booktitle}{\emph{Proceedings
  of the eleventh ACM international conference on web search and data mining}}.
  \bibinfo{pages}{565--573}.
\newblock


\bibitem[Van~den Oord et~al\mbox{.}(2013)]%
        {van2013deep}
\bibfield{author}{\bibinfo{person}{Aaron Van~den Oord}, \bibinfo{person}{Sander
  Dieleman}, {and} \bibinfo{person}{Benjamin Schrauwen}.}
  \bibinfo{year}{2013}\natexlab{}.
\newblock \showarticletitle{Deep content-based music recommendation}.
\newblock \bibinfo{journal}{\emph{Advances in neural information processing
  systems}}  \bibinfo{volume}{26} (\bibinfo{year}{2013}).
\newblock


\bibitem[Waite et~al\mbox{.}(2016)]%
        {waite2016project}
\bibfield{author}{\bibinfo{person}{Elliot Waite}, \bibinfo{person}{Douglas
  Eck}, \bibinfo{person}{Adam Roberts}, {and} \bibinfo{person}{Dan Abolafia}.}
  \bibinfo{year}{2016}\natexlab{}.
\newblock \showarticletitle{Project Magenta: Generating long-term structure in
  songs and stories}.
\newblock \bibinfo{journal}{\emph{Online] https://magenta. tensorflow.
  org/2016/07/15/lookback-rnn-attention-rnn}} (\bibinfo{year}{2016}).
\newblock


\bibitem[Wang and Wang(2014)]%
        {wang2014improving}
\bibfield{author}{\bibinfo{person}{Xinxi Wang} {and} \bibinfo{person}{Ye
  Wang}.} \bibinfo{year}{2014}\natexlab{}.
\newblock \showarticletitle{Improving content-based and hybrid music
  recommendation using deep learning}. In \bibinfo{booktitle}{\emph{Proceedings
  of the 22nd ACM international conference on Multimedia}}.
  \bibinfo{pages}{627--636}.
\newblock


\bibitem[Wang et~al\mbox{.}(2020)]%
        {wang2020pop909}
\bibfield{author}{\bibinfo{person}{Ziyu Wang}, \bibinfo{person}{Ke Chen},
  \bibinfo{person}{Junyan Jiang}, \bibinfo{person}{Yiyi Zhang},
  \bibinfo{person}{Maoran Xu}, \bibinfo{person}{Shuqi Dai},
  \bibinfo{person}{Xianbin Gu}, {and} \bibinfo{person}{Gus Xia}.}
  \bibinfo{year}{2020}\natexlab{}.
\newblock \showarticletitle{Pop909: A pop-song dataset for music arrangement
  generation}.
\newblock \bibinfo{journal}{\emph{arXiv preprint arXiv:2008.07142}}
  (\bibinfo{year}{2020}).
\newblock


\bibitem[Wu et~al\mbox{.}(2016)]%
        {wu2016collaborative}
\bibfield{author}{\bibinfo{person}{Yao Wu}, \bibinfo{person}{Christopher
  DuBois}, \bibinfo{person}{Alice~X Zheng}, {and} \bibinfo{person}{Martin
  Ester}.} \bibinfo{year}{2016}\natexlab{}.
\newblock \showarticletitle{Collaborative denoising auto-encoders for top-n
  recommender systems}. In \bibinfo{booktitle}{\emph{Proceedings of the ninth
  ACM international conference on web search and data mining}}.
  \bibinfo{pages}{153--162}.
\newblock


\bibitem[Xie et~al\mbox{.}(2022)]%
        {xie2022contrastive}
\bibfield{author}{\bibinfo{person}{Xu Xie}, \bibinfo{person}{Fei Sun},
  \bibinfo{person}{Zhaoyang Liu}, \bibinfo{person}{Shiwen Wu},
  \bibinfo{person}{Jinyang Gao}, \bibinfo{person}{Jiandong Zhang},
  \bibinfo{person}{Bolin Ding}, {and} \bibinfo{person}{Bin Cui}.}
  \bibinfo{year}{2022}\natexlab{}.
\newblock \showarticletitle{Contrastive learning for sequential
  recommendation}. In \bibinfo{booktitle}{\emph{2022 IEEE 38th international
  conference on data engineering (ICDE)}}. IEEE, \bibinfo{pages}{1259--1273}.
\newblock


\bibitem[Yang et~al\mbox{.}(2017)]%
        {yang2017midinet}
\bibfield{author}{\bibinfo{person}{Li-Chia Yang}, \bibinfo{person}{Szu-Yu
  Chou}, {and} \bibinfo{person}{Yi-Hsuan Yang}.}
  \bibinfo{year}{2017}\natexlab{}.
\newblock \showarticletitle{MidiNet: A convolutional generative adversarial
  network for symbolic-domain music generation}.
\newblock \bibinfo{journal}{\emph{arXiv preprint arXiv:1703.10847}}
  (\bibinfo{year}{2017}).
\newblock


\bibitem[Yi et~al\mbox{.}(2019)]%
        {yi2019singing}
\bibfield{author}{\bibinfo{person}{Yuan-Hao Yi}, \bibinfo{person}{Yang Ai},
  \bibinfo{person}{Zhen-Hua Ling}, {and} \bibinfo{person}{Li-Rong Dai}.}
  \bibinfo{year}{2019}\natexlab{}.
\newblock \showarticletitle{Singing voice synthesis using deep autoregressive
  neural networks for acoustic modeling}.
\newblock \bibinfo{journal}{\emph{arXiv preprint arXiv:1906.08977}}
  (\bibinfo{year}{2019}).
\newblock


\bibitem[Zhang et~al\mbox{.}(2022)]%
        {zhang2022structure}
\bibfield{author}{\bibinfo{person}{Xueyao Zhang}, \bibinfo{person}{Jinchao
  Zhang}, \bibinfo{person}{Yao Qiu}, \bibinfo{person}{Li Wang}, {and}
  \bibinfo{person}{Jie Zhou}.} \bibinfo{year}{2022}\natexlab{}.
\newblock \showarticletitle{Structure-enhanced pop music generation via
  harmony-aware learning}. In \bibinfo{booktitle}{\emph{Proceedings of the 30th
  ACM International Conference on Multimedia}}. \bibinfo{pages}{1204--1213}.
\newblock


\end{thebibliography}

\end{document}